\documentclass[]{article}
\usepackage{proceed2e}
\usepackage{times}
\usepackage{subfig} 
\usepackage{amsfonts}
\usepackage{amsmath}
\usepackage{amsthm}
\usepackage[psamsfonts]{amssymb}
\usepackage{nccmath}
\usepackage{mathtools} 
\usepackage{relsize} 
\usepackage{nicefrac}
\usepackage[square, numbers, authoryear]{natbib} 
\usepackage{stringstrings}
\usepackage{times}
\usepackage{bm}
\usepackage{bbm}
\usepackage{verbatim}
\usepackage{cancel}
\usepackage{verbatim}
\usepackage{color}
\usepackage{todonotes}
\usepackage{url}
\expandafter\def\expandafter\UrlBreaks\expandafter{\UrlBigBreaks
  \do\a\do\b\do\c\do\d\do\e\do\f\do\g\do\h\do\i\do\j%
  \do\k\do\l\do\m\do\n\do\o\do\p\do\q\do\r\do\s\do\t%
  \do\u\do\v\do\w\do\x\do\y\do\z\do\A\do\B\do\C\do\D%
  \do\E\do\F\do\G\do\H\do\I\do\J\do\K\do\L\do\M\do\N%
  \do\O\do\P\do\Q\do\R\do\S\do\T\do\U\do\V\do\W\do\X%
  \do\Y\do\Z\do\*\do\-\do\~\do\'\do\"\do\-\do\/}
\usepackage{epsfig}
\usepackage{mathtools}
\mathtoolsset{showonlyrefs}
\usepackage{tikz}
\usetikzlibrary{bayesnet}
\usetikzlibrary{positioning,patterns}
\makeatletter
\tikzset{nomorepostaction/.code=\let\tikz@postactions\pgfutil@empty}
\makeatother

\usetikzlibrary{fit}
\tikzset{%
  highlighth/.style={rectangle,rounded corners,fill=red!15,draw,
    fill opacity=0.3,thick,inner sep=0pt}
}
\tikzset{%
  highlightv/.style={rectangle,rounded corners,fill=green!15,draw,
    fill opacity=0.3,thick,inner sep=0pt}
}

\usepackage{mathdots}
\usepackage{algorithm}
\usepackage[noend]{algorithmic}
\usepackage{enumitem} 
\usepackage{eqparbox} 
\newlength\myindent
\newcommand{\alstatespace}{4pt}
\newcommand\SSTATE{\vspace{\alstatespace}\STATE}
\newcommand\SSTATEONE{\STATE}

\newcommand{\F}{\mathcal{F}}

\definecolor{light-gray}{gray}{0.43}

\global\long\def\outputIndex{j}
\global\long\def\dataIndex{i}

\global\long\def\latentIndex{j}

\global\long\def\dataDim{p}
\global\long\def\latentDim{q}

\global\long\def\numData{n}

\global\long\def\numInducing{m}

\global\long\def\dataScalar{y}
\global\long\def\dataVector{\mathbf{\dataScalar}}
\global\long\def\dataMatrix{\mathbf{\MakeUppercase{\dataScalar}}}

\global\long\def\latentScalar{x}
\global\long\def\latentMatrix{\mathbf{\MakeUppercase{\latentScalar}}}
\global\long\def\latentVector{\mathbf{\latentScalar}}

\global\long\def\kernelScalar{k}

\global\long\def\kernelMatrix{\mathbf{\MakeUppercase{\kernelScalar}}}

\global\long\def\numData{n}

\global\long\def\dataDim{p}

\global\long\def\mappingFunction{f}
\global\long\def\mappingFunctionVector{\mathbf{\mappingFunction}}
\global\long\def\mappingFunctionMatrix{\mathbf{\MakeUppercase{\mappingFunction}}}

\global\long\def\pseudotargetScalar{u}

\global\long\def\pseudotargetMatrix{\mathbf{\MakeUppercase{\pseudotargetScalar}}}

\global\long\def\inducingScalar{u}
\global\long\def\inducingVector{\mathbf{\inducingScalar}}
\global\long\def\inducingMatrix{\mathbf{\MakeUppercase{\inducingScalar}}}

\global\long\def\eye{\mathbf{I}}

\global\long\def\gaussianSamp#1#2{\mathcal{N}\left(#1,#2\right)}
\global\long\def\gaussianDist#1#2#3{\mathcal{N}\left(#1|#2,#3\right)}

\newcommand{\sparallel}{{\parallel}}
\global\long\def\KL#1#2{\text{KL}\left( #1\,\sparallel\,#2 \right)} 

\global\long\def\Kff{\kernelMatrix_{\mappingFunctionVector \mappingFunctionVector}}
\global\long\def\Kuu{\kernelMatrix_{\inducingVector \inducingVector}}

\global\long\def\Kfu{\kernelMatrix_{\mappingFunctionVector \inducingVector}}
\global\long\def\Kuf{\kernelMatrix_{\inducingVector \mappingFunctionVector}}

\global\long\def\tr#1{\text{tr}\left(#1\right)}

\usepackage{color}
\usepackage{verbatim}

\definecolor{brown}{rgb}{0.9,0.59,0.078}
\definecolor{ironsulf}{rgb}{0,0.7,.5}
\definecolor{lightpurple}{rgb}{0.156,0,0.245}

\ifdefined\blackBackground
\definecolor{colorOne}{rgb}{0, 1, 1}
\definecolor{colorTwo}{rgb}{1, 0, 1}
\definecolor{colorThree}{rgb}{1, 1, 0}
\definecolor{colorTwoThree}{rgb}{1, 0, 0}
\definecolor{colorOneThree}{rgb}{0, 1, 0}
\definecolor{colorOneTwo}{rgb}{0, 0, 1}
\else
\definecolor{colorOne}{rgb}{1, 0, 0}
\definecolor{colorTwo}{rgb}{0, 1, 0}
\definecolor{colorThree}{rgb}{0, 0, 1}
\definecolor{colorTwoThree}{rgb}{0, 1, 1}
\definecolor{colorOneThree}{rgb}{1, 0, 1}
\definecolor{colorOneTwo}{rgb}{1, 1, 0}
\fi

\ifdefined\blackBackground

\global\long\def\blackColor{white}
\global\long\def\whiteColor{black}
\else

\global\long\def\blackColor{black}
\global\long\def\whiteColor{white}
\fi

\global\long\def\bfmu{\boldsymbol{\mu}}

\global\long\def\bfepsilon{\boldsymbol{\epsilon}}
\global\long\def\bfSigma{\boldsymbol{\Sigma}}

\global\long\def\bftheta{\boldsymbol{\theta}}

\global\long\def\bff{\mathbf{f}}

\global\long\def\bft{\mathbf{t}}
\global\long\def\bfu{\mathbf{u}}

\global\long\def\bfy{\mathbf{y}}
\global\long\def\bfz{\mathbf{z}}

\global\long\def\bfzero{\mathbf{0}}

\global\long\def\eye{\mathbf{I}}
\global\long\def\bfK{\mathbf{K}}

\global\long\def\bfS{\mathbf{S}}

\global\long\def\bfU{\mathbf{U}}

\global\long\def\bfX{\mathbf{X}}
\global\long\def\bfY{\mathbf{Y}}
\global\long\def\bfZ{\mathbf{Z}}

\global\long\def\la{\leftarrow}

\global\long\def\T{{\top}}

\global\long\def\cut#1{}

\global\long\def\detail#1{}

\tikzstyle{obs} = [circle,inner sep=1pt,minimum size=20pt, font=\fontsize{10}{10}\selectfont, node distance=1,draw=\blackColor,fill=\blackColor!30]
\tikzstyle{latent} = [obs,fill=\whiteColor]
\tikzstyle{pixel} = [latent,minimum size=10pt, inner sep=0pt,node distance=0,font=\fontsize{5}{5}\selectfont]
\tikzstyle{constObserved} = [latent, node distance=1., fill=\blackColor!30, minimum size=5]
\tikzstyle{constUnobserved} = [latent, node distance=1., fill=\blackColor, minimum size=5]
\tikzstyle{dash plate} = [draw, rectangle, rounded corners, fit=#1]
\tikzstyle{dash plate caption} = [caption, node distance=0, inner sep=0pt,
above=5pt and 0pt of #1.north] %
\newcommand{\dashplate}[4][]{ %
  \node[wrap=#3] (#2-wrap) {}; %
  \node[dash plate caption=#2-wrap] (#2-caption) {#4}; %
  \node[dash plate=(#2-wrap)(#2-caption), #1] (#2) {}; %
}

\newcommand{\bfF}{\mathbf{F}}

\newcommand{\inducingIndex}{\dataIndex}

\newif\ifsubone
\subonetrue 

\newif\ifrowmatlabnotation          
\rowmatlabnotationtrue 

\newif\iflayersub                   
\layersubfalse 

\global\long\def\layerIndex{\ell}

\ifsubone
  \renewcommand{\dataIndex}{i}        
  \renewcommand{\outputIndex}{j}      
  \renewcommand{\latentIndex}{j}      
  \renewcommand{\inducingIndex}{i}    
\fi

\renewcommand{\numData}{n}            
\renewcommand{\dataDim}{p}            
\renewcommand{\latentDim}{q}          
\renewcommand{\numInducing}{m}        

\newcommand{\jn}{\dataIndex}
\newcommand{\jd}{\outputIndex}
\newcommand{\jq}{\latentIndex}

\newcommand{\jm}{\inducingIndex}

\newcommand{\sxo}{z}
\newcommand{\vxo}{\bfz}
\newcommand{\mXo}{\bfZ}

\newcommand{\ie}{i.e.\ }
\newcommand{\eg}{e.g.\ }

\newcommand{\ra}{\right\rangle}

\newcommand{\intd}{\text{d}}
\renewcommand{\t}[1]{(#1)^\T}

\newcommand{\observedSet}{{\mathsmaller{\mathcal{O}}}}
\newcommand{\unobservedSet}{{\mathsmaller{\mathcal{U}}}}
\newcommand{\labelledSet}{{\mathsmaller{\mathcal{L}}}}
\newcommand{\unlabelledSet}{{\mathsmaller{\mathcal{M}}}}

\renewcommand{\la}{\left\langle}

\ifsubone
  \newcommand{\tn}{_\dataIndex}  
  \newcommand{\n}{_\dataIndex}                
  \renewcommand{\d}{_\outputIndex}            
  \newcommand{\nd}{_{\dataIndex,\outputIndex}}    
  \newcommand{\q}{_\latentIndex}              
  \newcommand{\nq}{_{\dataIndex,\latentIndex}}    
  \newcommand{\m}{_\inducingIndex}

  \newcommand{\inn}[1]{_{#1,:}}                
  \newcommand{\inq}[2]{_{#1,#2}}               
  \renewcommand{\l}{^{(\layerIndex)}}             

  \newcommand{\innTs}[1]{_{#1,*}}                
  \ifrowmatlabnotation
    \renewcommand{\n}{_{\dataIndex,:}}                
    \renewcommand{\m}{_{\inducingIndex,:}}
  \fi
\else
  \newcommand{\n}{_{\dataIndex,:}}                
  \renewcommand{\d}{_{:,\outputIndex}}            
  \newcommand{\nd}{_{\dataIndex,\outputIndex}}    
  \newcommand{\q}{_{:,\latentIndex}}              
  \newcommand{\nq}{_{\dataIndex,\latentIndex}}    
  \newcommand{\m}{_{\inducingIndex, :}}

  \newcommand{\inn}[1]{_{#1,:}}                
  
  \newcommand{\inq}[2]{_{#1,#2}}               
  \renewcommand{\l}{^{(\layerIndex)}}             

  \newcommand{\innTs}[1]{_{#1,*}}                
\fi

\newcommand{\nN}{\numData}
\newcommand{\nQ}{\latentDim}
\newcommand{\nD}{\dataDim}

\newcommand{\sy}{\dataScalar}
\renewcommand{\sf}{\mappingFunction} 
\newcommand{\su}{\pseudotargetScalar}

\newcommand{\vx}{\latentVector}
\newcommand{\vy}{\dataVector}
\newcommand{\vf}{\mappingFunctionVector}

\newcommand{\mX}{\latentMatrix}
\newcommand{\mY}{\dataMatrix}
\newcommand{\mF}{\mappingFunctionMatrix}

\newcommand{\mZ}{\latentMatrix_\su}
\newcommand{\mU}{\pseudotargetMatrix}

\newcommand{\xn}{\vx\n} 
\newcommand{\xq}{\vx\q} 

\newcommand{\xon}{\vxo\n} 
\newcommand{\xonq}{\sxo\nq}

\newcommand{\yn}{\vy\n} 
\newcommand{\yd}{\vy\d} 
\newcommand{\ynd}{\sy\nd}

\newcommand{\fd}{\vf\d} 

\renewcommand{\Kff}{\bfK}

\title{Semi-described and semi-supervised learning with Gaussian processes}

\author{
{\bf Andreas Damianou}\\ 
Dept. of Computer Science \& SITraN \\
The University of Sheffield \\
Sheffield, UK
\And
{\bf Neil D. Lawrence}\\
Dept. of Computer Science \& SITraN  \\
The University of Sheffield \\
Sheffield, UK
} 

\begin{document}

\maketitle

\begin{abstract}
Propagating input uncertainty through non-linear Gaussian process (GP) mappings is intractable.
This hinders the task of training GPs using uncertain and partially observed inputs. In this paper we refer to this task as ``semi-described learning''. We then introduce a GP framework that solves both, the semi-described and the semi-supervised learning problems (where missing values occur in the \emph{outputs}). Auto-regressive state space simulation is also recognised as a special case of semi-described learning. To achieve our goal we develop variational methods for handling semi-described inputs in GPs, and couple them with algorithms that allow for imputing the missing values while treating the uncertainty in a principled, Bayesian manner.
Extensive experiments on simulated and real-world data study the problems of iterative forecasting and regression/classification with missing values.
The results suggest that the principled propagation of uncertainty stemming from our
framework can significantly improve performance in these tasks.
\end{abstract}

\section{INTRODUCTION\label{sec:intro}}

In many real-world applications missing values can occur in the data, for example when measurements come from unreliable sensors. Correctly accounting for the partially observed instances is important in order to exploit all available information and increase the strength of the inference model. The focus of this paper is on Gaussian process (GP) models that allow for Bayesian, non-parametric inference.
 
When the missing values occur in the outputs, the corresponding learning task is known as \emph{semi-supervised learning}. 
For example, consider the task of learning to classify images where the labelled set is much smaller than the total set. Bootstrapping is a potential solution to this problem \citep{Rosenberg:semiSupervisedSelfTraining}, according to which a model trained on fully observed data imputes the missing outputs. Previous work in semi-supervised GP learning involved the cluster assumption \citep{Lawrence:semisuper04} for classification. Here we consider an approach which uses the manifold assumption \citep{Chapelle:semisuper06,Kingma14:semiSupervisedDeepGenerative} which assumes that the observed, complex data are really generated by a compressed, less-noisy latent space.

The other often encountered missing data problem has to do with unobserved \emph{input} features (\eg missing pixels in input images).  In statistics, a popular approach is to impute missing inputs using a combination of different educated guesses \citep{rubin2004multiple}. In machine learning, \citet{Ghahramani:missing94} learn the joint density of the input and output data and integrate over the missing values. For Gaussian process models 
the missing input case has received only little attention,
due to the challenge of propagating the input uncertainty through the non-linear GP mapping. In this paper we introduce the notion of \emph{semi-described learning} to generalise this scenario. Specifically, we define semi-described learning to be the task of learning from inputs that can have missing or uncertain values.
Our approach to dealing with missing inputs in semi-described GP learning is, algorithmically, closer to data imputation methods. However, in contrast to past approaches, the missing values are imputed in a fully probabilistic manner by considering explicit distributions in the input space. 

Our aim in this paper is to develop a general framework that solves the semi-supervised and semi-described GP learning. We also consider the related \emph{forecasting regression} problem, which is seen as a pipeline where predictions are obtained iteratively in an auto-regressive manner, while propagating the uncertainty across the predictive sequence, as in \citep{Girard:uncertain01Compact,quinonero2003propagation}. Here, we cast the auto-regressive GP learning as a particular type of semi-described learning. We seek to solve all tasks within a single coherent framework that preserves the fully Bayesian  property of the GP methodology.

To achieve our goals we need three methodological tools. Firstly, we need approximations allowing us to consider and communicate uncertainty between the inputs and the outputs of the non-linear GP model. 
For this, we build on the variational approach of \cite{BayesianGPLVM} which allows for approximately propagating densities throughout the nodes of GP-based directed
graphical models. The resulting representation is particularly advantageous, because the whole input domain is now coherently associated with posterior distributions. We can then sample from the input space in a principled manner so as to populate small initial labelled sets in semi-supervised learning scenarios. In that way, we avoid heuristic self-training methods \citep{Rosenberg:semiSupervisedSelfTraining} that rely on boot-strapping and present problems due to over-confidence. Previously suggested approaches for modelling input uncertainty in GPs also lack the feature of considering an explicit input distribution for both training and test instances. Specifically, \citep{Girard:uncertain01Compact,quinonero2003propagation} consider the case of input uncertainty \emph{only at test time}. Propagating the test input uncertainty through a non-linear GP results in a non-Gaussian predictive density, but \cite{Girard:uncertain01Compact,quinonero2003propagation,quinonero2004learning} rely on moment matching to obtain the predictive mean and covariance. On the other hand, \citet{Oakley:computer02} do not derive analytic expressions but, rather, develop a scheme based on simulations. \citet{mchutchon:gaussian} rely on local approximations inside the latent mapping function, rather than modelling the approximate posterior densities directly. \cite{dallaire2009learning} do not propagate the uncertainty of the inputs all the way through the GP mapping but, rather, amend the kernel computations to account for the input uncertainty. \citep{Candela:clearning03} can be seen as a special case of our developed framework, when the data imputation is performed using a standard GP-LVM \citep{Lawrence:gplvmtut06}.
Another advantage of our framework is that it allows us to consider different levels of input uncertainty per point and per dimension without, in principle, increasing the danger of under/overfitting, since input uncertainty is modelled through a set of \emph{variational} rather than model parameters.

The second methodological tool needed to achieve our goals has to do with the need to incorporate partial or uncertain observations into the variational framework.
For this, we develop a \emph{variational constraint} mechanism which constrains the distribution of the input space given the observed noisy values. 
 This approach is fast, and the whole framework can be incorporated into a parallel inference algorithm \citep{Gal:Distributed14,Dai2014Parallel}.
In contrast, \cite{Damianou:vgpds11Compact} consider a separate process for modelling the input distribution. However, that approach cannot easily be extended for the data imputation purposes that concern us, since we cannot consider different uncertainty levels per input and per dimension and, additionally, computation scales cubicly with the number of datapoints, even within sparse GP frameworks. 
The constraints framework that we propose is interesting not only as an inference tool but also as a modelling approach: if the inputs are constrained with the outputs, then we obtain the Bayesian version of the back-constraints framework of  \cite{Lawrence:backconstraints06} and \cite{Ek:ambiguity08}. However, 
in contrast to these approaches, the constraint defined here is a \emph{variational} one, and operates upon a distribution, rather than single points. 
\cite{zhu2012medlda} also follow the idea of constraining the posterior distribution with rich side information, albeit for a completely different application. In contrast, \cite{Osborne:gpreport2007} handle partially missing sensor inputs by modelling correlations in the input space through special covariance functions.

Thirdly, the variational methods developed here need to be encapsulated into algorithms that perform data imputation while correctly accounting for the introduced uncertainty. We develop such algorithms after  showing how the considered applications can be cast as learning pipelines that rely on correct propagation of uncertainty between each stage.

In summary, our contributions in this paper are the following; firstly, by building on the Bayesian GP-LVM \citep{BayesianGPLVM} and developing a variational constraint mechanism we demonstrate how uncertain GP inputs can be explicitly represented as distributions in both training and test time. Secondly, we couple our variational methodology with algorithms that allow us to solve problems associated with partial or uncertain observations: semi-supervised learning, auto-regressive iterative forecasting and, finally, a newly studied type of GP learning which we refer to as ``semi-described'' learning. We solve these applications within a single framework, allowing for handling the uncertainty in semi-supervised and semi-described problems in a coherent way. The software accompanying this paper can be found at: \url{http://git.io/A3TN}. This paper extends our previous workshop paper \citep{Damianou:pipelines14}.

\section{UNCERTAIN INPUTS REFORMULATION OF GP MODELS \label{sec:unc_model}}

Assume a dataset of input--output pairs stored by rows
in matrices $\mX \in \Re^{\nN \times \nQ}$ and
$\mY \in \Re^{\nN \times \dataDim}$
respectively. 
 Throughout this paper we will denote rows of the above matrices as $\{\yn, \xn \}$ and columns (dimensions) as $\{ \yd, \xq\}$, while single elements (\eg $\ynd$) will be denoted with a double subscript. 
We first outline the standard GP formulation, where all variables are fully observed. 
  By assuming that outputs are corrupted by zero-mean Gaussian noise, denoted by $\bfepsilon_f$, we obtain the following generative model:
\begin{align}
\ynd &= \fd (\xn) + (\epsilon_f)\nd,  \quad  (\epsilon_f)\nd  \sim \gaussianSamp{0}{\beta^{-1}} .
\label{eq:generative1}
\end{align}
We place GP priors on the mapping $\vf$, so that the function instantiations $\mF = \{ \fd \}_{\jd=1}^\nD$ follow a Gaussian distribution
$p(\fd | \mX) = \gaussianDist{\fd}{\bfzero}{\Kff}$,
where $\Kff$ is the covariance matrix obtained by evaluating the GP covariance function $\kernelScalar_\sf$ on the inputs $\mX$.
Therefore, the model likelihood $p(\mY | \mX)$ is:
\begin{equation}
\label{eq:likelihood}
\int_\mF p(\mY | \mF) p(\mF | \mX) 
= \prod_{\jd=1}^{\nD} \gaussianDist{\yd}{\bfzero}{\Kff + \beta^{-1}\eye}.
\end{equation}

In the other end of the spectrum is the GP-LVM \citep{Lawrence:gplvmtut06}, where the inputs are fully unobserved (\ie \emph{latent}). This corresponds to the unsupervised GP setting. In the absence of observed inputs, the likelihood $p(\mY | \mX)$ takes the same form as in equation \eqref{eq:likelihood} but the inputs $\mX$ now need to be recovered from the outputs $\mY$ through maximum likelihood.
The Bayesian GP-LVM proceeds by additionally placing a Gaussian prior on the latent space, $p(\mX) = \prod_{\jn=1}^\nN \gaussianDist{\xn}{\bfzero}{\eye}$, and approximately integrating it out
by constructing a variational lower bound $\F$, where
\begin{equation}
\label{eq:bgplvmMarginalLikelihood}
\F \le \log p(\mY) = \log \int_\mX p(\mY | \mX) p(\mX),
\end{equation}
and by introducing a variational distribution
\begin{equation}
\label{eq:posterior}
q(\mX) = \prod_{\jn=1}\nolimits^\nN q(\xn) = \prod_{\jn=1}\nolimits^\nN \gaussianDist{\xn}{\bfmu\n}{\bfS\n},
\end{equation} 
where $\bfS\n$ is a diagonal matrix, so that $\bfmu\n,\text{diag}(\bfS\n) \in \Re^\nQ$.
We can derive an expression for this variational bound,
\begin{equation}
\label{eq:bound}
\F = \la \log p(\mY|\mX)\ra_{q(\mX)} - 
\KL{q(\mX)}{p(\mX)},
\end{equation} 
where $\la \cdot \ra_{q(\mX)}$ denotes an expectation with respect to $q(\mX)$. 
Since $\mX$ appears non-linearly inside $p(\mY | \mX)$ (in the inverse of the covariance matrix $\Kff + \beta^{-1} \eye$), the first term of the above variational bound is intractable.
However, we can follow \citep{BayesianGPLVM} to approximate the intractable expectation analytically.

In this paper we wish to define a general framework that operates in the whole range of the two aforementioned extrema, \ie the fully observed and fully unobserved inputs case. The first step to obtaining such a framework is to allow for uncertainty in the inputs. We assume that the inputs $\mX$ are not observed directly but, rather, we only have access to their noisy versions $ \{ \xon \}_{\jn=1}^\nN = \mXo \in \Re^{\nN \times \nQ}$. The relationship between the noisy and true inputs is given by assuming Gaussian noise:
\begin{align}
\xn &= \xon + (\bfepsilon_x)\n,  \quad (\bfepsilon_x)\n \sim \gaussianSamp{\bfzero}{\bfSigma_x} 
\label{eq:generative2},
\end{align}
so that $p(\mX | \mXo) = \prod_{\jn=1}^\nN \gaussianDist{\xn}{\xon}{\bfSigma_x}$.
Obviously, when this distribution collapses to a delta function we recover the standard GP case, and when $\mXo$ is not given we recover the GP-LVM.
The problem with the modelling assumption of equation \eqref{eq:generative2} is that now we cannot use equation \eqref{eq:generative1}, since the inputs are not available. On the other hand, if we replace $\xn$ in that equation with $\xon$, then we effectively ignore the input noise. \citet{mchutchon:gaussian} proceed by combining equations \eqref{eq:generative1} and \eqref{eq:generative2} to obtain the GP mapping $\fd(\xn - (\bfepsilon_x)\n )$ which is then treated using local approximations.
However, our aim in this paper is to consider an explicit input distribution. One way to achieve this is to treat the unobserved true inputs as latent variables to be estimated from the marginal likelihood $p(\mY|\mXo) = \int_\mX p(\mY|\mX)p(\mX|\mXo)$. Following \citet{Damianou:vgpds11Compact} we can obtain a variational lower bound $\F \le \log p(\mY|\bfZ)$, with:

\begin{equation}
\label{eq:bound2}
\F = \la \log p(\mY|\mX)\ra_{q(\mX)} - 
\KL{q(\mX)}{p(\mX | \bfZ)} .
\end{equation} 

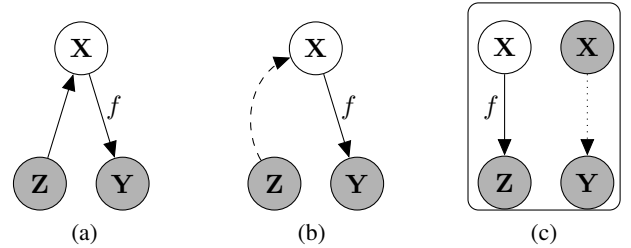
\begin{figure}[ht]
\def\layersep{1.8cm}
\def\nodesep{0cm}
\def\nodepos{0.55}
\begin{center}
\subfloat[]{  
    \begin{tikzpicture}[node distance=\layersep, scale=1, every node/.style={transform shape},line/.style={->,dashed}]
      \tikzstyle{annot} = [text width=4em, text centered]    
        \node[obs] (Z) at (-\nodepos, 0) {$\color{black}{\bfZ}$};
        \node[obs] (Y) at (\nodepos, 0) {$\color{black}{\mY}$};
        \path[xshift=0*\nodesep]
          node[latent] (X) at (0, \layersep) {$\color{black}{\mX}$};
      \draw[->] (X) -- node[pos=0.4] {$\; \; \; \; f$} (Y);
      \draw[->] (Z) -- (X);
    \end{tikzpicture}
    \label{subfig:model1}
}%
\hfill
\subfloat[]{
    \begin{tikzpicture}[node distance=\layersep, scale=1, every node/.style={transform shape},line/.style={->,dashed}]
      \tikzstyle{annot} = [text width=4em, text centered]    
        \node[obs] (Z) at (-\nodepos, 0) {$\color{black}{\bfZ}$};
        \node[obs] (Y) at (\nodepos, 0) {$\color{black}{\mY}$};
        \path[xshift=0*\nodesep]
          node[latent] (X) at (0, \layersep) {$\color{black}{\mX}$};
      \draw[->] (X) -- node[pos=0.4] {$\; \; \; \; f$} (Y);
      \path [black,line,out=120,in=200] (Z) edge (X); 
    \end{tikzpicture}
    \label{subfig:model2}
}%
\hfill
\subfloat[]{
    \begin{tikzpicture}[node distance=\layersep, scale=1, every node/.style={transform shape},line/.style={->,dotted}]
      \tikzstyle{annot} = [text width=4em, text centered]    
        \node[obs] (Y) at (-\nodepos, 0) {$\color{black}{\bfZ}$};
        \node[obs] (L) at (\nodepos, 0) {$\color{black}{\mY}$};
        \node[] (f) at (-0.73, \layersep/1.8) {$f$};
        \path[xshift=0*\nodesep]
          node[latent] (X1) at (-\nodepos, \layersep) {$\color{black}{\mX}$};
        \path[xshift=0*\nodesep]
          node[obs] (X2) at (\nodepos, \layersep) {$\color{black}{\mX}$};
      \draw[->] (X1) -- (Y);
      \path [black,line] (X2) edge (L);
      \dashplate[inner xsep=2pt, inner ysep=0pt] {plate} {(Y)(L)(X1)(X2)(f)} {}
    \end{tikzpicture}
    \label{subfig:model3}
}
\end{center}
\vspace{-10pt}
\caption{
Incorporating uncertain inputs $\bfZ$ in GPs through an intermediate input space $\mX$ by considering: \subref{subfig:model1} a Gaussian prior on $\mX$, centered on $\bfZ$ and \subref{subfig:model2} a  variational constraint (dashed line) on the approximate posterior. Figure \subref{subfig:model3} represents our two-stage approach to dealing with missing outputs for classification, where the dotted line represents a discriminative mapping.}
\label{fig:models}
\end{figure}

This formulation corresponds to the graphical model of Figure
\ref{fig:models}\subref{subfig:model1}.  However, with this approach
one needs to additionally estimate the noise parameters $\bfSigma_x$,
which might be challenging given their large number and their
interplay with the variational noise parameters $\{ \bfS\n
\}_{\jn=1}^\nN$.  Therefore we considered
an alternative solution which we found to result in 
better performance.  

\subsection{VARIATIONAL CONSTRAINT\label{sec:backConstraint}}

An alternative way of relating the true with the noisy inputs can be obtained by focusing on the \emph{posterior} rather than the prior distribution. To start with, we re-express the variational lower bound of equation \eqref{eq:bound2} as:
\begin{align*}
\log p(\mY | \mXo) 
& \ge \int_\mX q(\mX) \log 
                          \frac{p(\mY |\mXo) p(\mX | \mY, \mXo)}
                               {q(\mX)} 
    = \F
\end{align*}
from where we break the logarithm to obtain:
\begin{align}
\F = \log p(\mY | \mXo) - \KL{q(\mX)}{p(\mX | \mY, \mXo)}.
\end{align}
We see that the lower bound becomes exact when the variational
distribution $q(\mX)$ matches the true posterior
distribution of the noise-free latent inputs given the observed inputs and
outputs. 
To allow for this approximation we introduce a simple
variational constraint which operates on the factorised
distribution, which is now written as $q(\mX | \bfZ)$ to
highlight its dependency on $\bfZ$. In the simplest case where all inputs are observed but uncertain, the constraint just
consists of replacing the variational means $\bfmu\n$ of each
factor $q(\xn)$ with the corresponding observed
input $\xon$. The variational parameters $\bfS\n$ then account for the uncertainty. Similarly to the back-constraint of
\citet{Lawrence:backconstraints06,Ek:ambiguity08}, our variational 
constraint does not constitute a probabilistic mapping. 
However, it allows us to
encode the input noise directly in the approximate posterior without
having to specify additional noise parameters or sacrifice
scalability. Next, we elaborate on the exact form of the constraint.

In the general case, namely having inputs that are only partially observed, we can define a similar constraint which specifies a variational distribution as a mix of Gaussian and Dirac delta distributions. 
Notationally we consider data to be split into fully and  partially observed subsets, \eg $\bfZ = (\bfZ^\observedSet, \bfZ^\unobservedSet)$, where $\observedSet$ and $\unobservedSet$ denote fully and partially observed sets respectively.
The features missing in $\bfZ^\unobservedSet$ can appear in different dimension(s) for each individual point $\xon^\unobservedSet$, but for notational clarity $\unobservedSet$ will index rows containing at least one missing dimension. In this case, the variational distribution is constrained to have the form
\begin{align}
&q(\mX | \bfZ,  \{ \mathsmaller{\observedSet}, \mathsmaller{\unobservedSet} \}) 
= q(\mX^\observedSet | \bfZ^\observedSet) \; q(\mX^\unobservedSet | \bfZ^\unobservedSet) \nonumber  \\
&= \prod\nolimits_{\jn \in \observedSet} \gaussianDist{\xn^\observedSet}
                                                {\xon^\observedSet}
                                                {\varepsilon \eye} 
 \prod\nolimits_{\jn \in \unobservedSet} \gaussianDist{\xn^\unobservedSet}
                                                {\bfmu\n^\unobservedSet}
                                                {\bfS\n^\unobservedSet}, \; \; \;   \label{eq:posteriorSemiSupervised}
\end{align}
where $\varepsilon \rightarrow 0$, so that the corresponding distributions approximate a Dirac delta.
Notice that for a partially observed row $\xon^\unobservedSet$, we can still replace an observed dimension $\latentIndex$ with its corresponding observation in the second set of factors of equation \eqref{eq:posteriorSemiSupervised}, \ie $\mu_{\jn,\latentIndex}^\unobservedSet = z_{\jn,\latentIndex}^\unobservedSet$,
 so  $q(\mX^\unobservedSet | \bfZ^\unobservedSet) \ne q(\mX^\unobservedSet)$.
Given the above, as well as a spherical Gaussian prior for $p(\mX)$, the required intractable density $\log p(\mY | \bfZ)$ is approximated with a variational lower bound:
\begin{equation}
\label{eq:boundSemiSupervised}
\F = \la \log p(\mY|\mX)\ra_{q(\mX | \bfZ)} - 
\KL{q(\mX | \bfZ)}{p(\mX)} ,
\end{equation}
where for clarity we dropped the dependency on $\{ \mathsmaller{\observedSet},\mathsmaller{\unobservedSet} \}$ from our expressions.
Since the Dirac functions are approximated with sharply peaked Gaussians inside the posterior $q(\mX | \bfZ)$, the above variational bound can be computed in the same manner as the Bayesian GP-LVM bound of equation \eqref{eq:bound}. Specifically, the KL term is tractable, since it only involves Gaussians.

As for the first term of equation \eqref{eq:boundSemiSupervised}, we follow the Bayesian GP-LVM methodology and  we augment the probability space with $\numInducing$ extra samples $\inducingMatrix = \{ \inducingVector_\jm \}_{\jm=1}^\numInducing$ of the latent function $\mappingFunction$ evaluated at a set of pseudo-inputs (known as ``inducing points'') $\mX_\inducingScalar$, so that $\inducingMatrix \in \Re^{\numInducing \times \dataDim}$ and $\mX_\inducingScalar \in \Re^{\numInducing \times \nQ}$. Due to the consistency of GPs, $p(\inducingMatrix | \mX_\inducingScalar)$ is a Gaussian distribution. From now on we omit dependence on $\mZ$ from our expressions. The likelihood then becomes:
\begin{equation}
\label{eq:augmentedLikelihood}
p(\mY,\mF,\mU | \mX) = p(\mY|\mF) p(\mF|\mU,\mX) p(\mU).
\end{equation}
Then, the marginal $p(\mY|\mX)$ can be obtained from Jensen's inequality after introducing a variational distribution $q(\mF,\mU)$, so that $\hat{\F} \le \log p(\mY | \mX)$, where:
\begin{equation}
\label{eq:jensens2}
\hat{\F} = \int_{\mF,\mU} q(\mF,\mU) \log \frac{p(\mY|\mF) p(\mF|\mU,\mX) p(\mU)}{q(\mF,\mU)}.
\end{equation}
Now the fist term of equation \eqref{eq:boundSemiSupervised} is approxmated as  $\la p(\mY|\mX) \ra_{q(\mX | \mXo)} \ge \small\langle \hat{\F} \small\rangle_{q(\mX|\mXo)}$. However, this approximation is still intractable, since the problematic term $p(\mF|\mU,\mX)$ still appears inside $\hat{\F}$ and contains $\mX$ in the inverse of the covariance matrix, thus rendering the expectation intractable. The trick of \citet{BayesianGPLVM} is to define a variational distribution of the form:
\begin{equation}
\label{eq:varDistr}
q(\mF,\mU) = p(\mF|\mU,\mX) q(\mU). 
\end{equation}
Replacing equation \eqref{eq:varDistr}  inside the bound of equation \eqref{eq:jensens2} results in the cancellation of $p(\mF|\mU,\mX)$, leaving us with a tractable (partial) bound, which takes the form:
\begin{align}
\small
& \la p(\mY|\mX) \ra_{q(\mX|\mXo)} \ge \langle \hat{\F} \rangle_{q(\mX|\mXo)} =  - \KL{q(\inducingMatrix)}{p(\inducingMatrix )}  \nonumber \\
& + \int_{\mX, \inducingMatrix} \left[ q(\mX | \bfZ) q(\inducingMatrix) \int_\mF p(\mF | \inducingMatrix, \mX) \log p(\mY | \mF) \right] \label{eq:boundFinal}. 
\end{align}
 The augmentation trick decouples the latent function values given the inducing points, so that any uncertainty in the inputs can be propagated through the nested integral. After this operation, the inducing outputs $\inducingMatrix$ can be marginalised out. Therefore, the above integral is analytically tractable, since the nested integral is tractable and results in a Gaussian where $\mX$ no longer appears in the inverse of the covariance matrix. The final lower bound to use as an objective function is thus obtained by using the partial bound of eq. \eqref{eq:boundFinal} in place of the first term of equation \eqref{eq:boundSemiSupervised}, thus obtaining a new, final bound (more details in the Appendix):
 \begin{equation}
 \label{eq:boundFinalFinal}
 \F_2 = \small\langle \hat{\F} \small\rangle_{q(\mX|\mXo)} - \KL{q(\mX | \bfZ)}{p(\mX)}.
 \end{equation}

To summarise, the variational methodology seeks to approximate the true posterior with a variational distribution $q(\mF,\mU,\mX)=$ $q(\mF)q(\mU)q(\mX)$. To achieve this, $q(\mF)$ is constrained to take the exact form $p(\mF|\mU,\mX)$. This term is then ``eliminated'', giving us tractability, but its effect is re-introduced through the variational distribution (in the nested integral of eq. \eqref{eq:boundFinal}). Contrast this with the variational constraint on $q(\mX)$: that approximate posterior factor is constrained according to $\mXo$, so that the effect of $\mXo$ is considered only through the $q(\mX|\mXo)$  (eq. \eqref{eq:boundFinalFinal}). The above comparison gives insight in the conceptual similarity of the variational approach followed to obtain tractability and the one followed for handling partially observed inputs.

The variationally constrained model is shown in fig. 
\ref{fig:models}\subref{subfig:model2}. The total set of parameters
to be optimised in the objective function $\F_2$ of equation \eqref{eq:boundFinalFinal} (\eg using a gradient-based optimiser)
 are the model parameters $(\bftheta_f,
\beta)$, where $\bftheta_f$ denotes the hyper-parameters of the
covariance function $k_\mappingFunction$, and the variational
parameters $(\mX_\inducingScalar, \{
\bfmu\n^\unobservedSet, \bfS\n^\unobservedSet
\}_{\jn \in \unobservedSet})$ 
($q(\bfU)$ can be optimally eliminated, see Appendix).
 Depending on the application and
corresponding learning algorithm, certain dimensions of $\{
\bfmu\n^\unobservedSet, \bfS\n^\unobservedSet \}$
can be treated as observed. Such algorithms are discussed in the
following sections.

\section{GP LEARNING WITH MISSING VALUES\label{sec:missingValuesLearning}}
We formulate both the semi-described and semi-supervised learning as particular instances of learning a mapping function where the inputs are associated with uncertainty. 
In both cases, we devise a two-step strategy based on our uncertain inputs GP framework, which allows to efficiently take into account the partial information in the given datasets to improve the predictive performance. For brevity, we refer to the framework described in the previous section as a \emph{variationally constrained GP}, from where a semi-described, an auto-regressive and a semi-supervised GP approach are obtained as special cases, given the algorithms that will be explained in this section.

\subsection{SEMI-DESCRIBED LEARNING\label{subsec:semiDescribed}}
We assume a set of observed outputs $\mY$ that correspond to fully observed inputs $\mXo^\observedSet$ and partially observed inputs $\mXo^\unobservedSet$, so that $\mXo = (\mXo^\observedSet, \mXo^\unobservedSet)$. To make the correspondence clearer, we also split the observed \emph{outputs} according to the sets $\{ \observedSet, \unobservedSet \}$, so that $\mY = (\mY^\observedSet, \mY^\unobservedSet)$, but note that both output sets are fully observed. We are then interested in learning a regression function from $\mXo$ to $\mY$ by using all available information.
Since in the variationally constrained GP the inputs are replaced by distributions $q(\mX^\observedSet | \mXo^\observedSet)$ and $q(\mX^\unobservedSet | \mXo^\unobservedSet)$, the uncertainty over $\mXo^\unobservedSet$ can be taken into account naturally through this variational distribution. In this context, we formulate a data imputation-based approach which is inspired by self-training methods; nevertheless, it is more principled in the handling of uncertainty. 

Specifically, the algorithm has two stages; in the first step,
 we use the fully observed data subset $(\mXo^\observedSet, \mY^\observedSet)$ to train an initial variationally constrained GP model which encapsulates the sharply peaked variational distribution $q(\mX^\observedSet | \mXo^\observedSet)$ given in equation \eqref{eq:posteriorSemiSupervised}. Given this model, we can then use $\mY^\unobservedSet$ to estimate the predictive posterior\footnote{The predictive posterior for test data $\mY_*$ is obtained by maximising a variational lower bound similar to the training one (eq. \eqref{eq:boundFinalFinal}), but $\mX$ and $\mY$ are now replaced with $(\mX,\mX_*)$ and $(\mY,\mY_*)$.} $q(\mX^\unobservedSet | \mXo^\unobservedSet$) in the missing locations of $\mXo^\unobservedSet$ (for the observed locations we match the mean with the observations in a sharply peaked marginal, as for $\mXo^\observedSet$).
Essentially, we replace the missing locations of the variational means $\bfmu\n^\unobservedSet$ and variances $\bfS\tn^\unobservedSet$ of $q(\mX^\unobservedSet | \mXo^\unobservedSet)$ with the predictive mean and variance obtained through the ``self-training'' step. This selection for $\{ \bfmu\n^\unobservedSet, \bfS\tn^\unobservedSet \}$ constitutes nevertheless only an initialisation. 
In the next step, these parameters are further optimised together with the fully observed data.
Specifically, 
 after initializing $q(\mX |\mXo)=q(\mX^\observedSet, \mX^\unobservedSet | \mXo)$ as explained in step 1, we proceed to train a variationally constrained GP model on the full (extended) training set
$\left( \left( \mXo^\observedSet, \mXo^\unobservedSet \right),
        \left( \mY^\observedSet, \mY^\unobservedSet \right)
\right)$,
which contains fully and partially observed inputs.

\begin{algorithm*}[t]
\small
\caption[Semi-described learning with GPs.]{Semi-described learning with uncertain input GPs.}\label{algorithm:semiSupervised}
\begin{algorithmic}[1]
\SSTATEONE \emph{Given}: Fully and partially observed inputs, $\mXo^\observedSet$ and $\mXo^\unobservedSet$ respectively, corresponding  to fully observed outputs $\mY^\observedSet$ and $\mY^\unobservedSet$.
\SSTATE \label{state:qObs} Construct $q(\mX^\observedSet | \mXo^\observedSet) = \prod_{\jn=1}^\nN
      \gaussianDist{\xn^\observedSet}
                    {\xon^\observedSet}
                    {\varepsilon \eye}, \text{where: } \varepsilon \rightarrow 0$
\SSTATE Fix $q(\mX^\observedSet | \mXo^\observedSet)$ in the optimiser \COMMENT{\emph{(\ie $q(\mX^\observedSet | \mXo^\observedSet)$ has no free parameters)}}
\SSTATE Train a variationally constrained GP model $\mathcal{M}^\observedSet$ with inputs $q(\mX^\observedSet | \mXo^\observedSet )$ and outputs $\mY^\observedSet$
\vspace{\alstatespace}
\FOR{$\jn = 1, \cdots, |\mY^\unobservedSet|$}\label{state:forStart}
    \SSTATEONE Predict the distribution $\gaussianDist{\vx\n^\unobservedSet}{\hat{\bfmu}^{\unobservedSet}\n}{\hat{\bfS}^{\unobservedSet}\tn}
    \approx 
    p(\vx\n^\unobservedSet |  \yn^\unobservedSet, \mathcal{M}^\observedSet) 
    $ from the approximate posterior of model $\mathcal{M}^\observedSet$. 
    \SSTATE Initialise parameters $\{\bfmu\n^\unobservedSet,\bfS\tn^\unobservedSet\}$ of $q(\xn^\unobservedSet | \xon^\unobservedSet) = 
    \gaussianDist{\xn^\unobservedSet}
                 {\bfmu\n^\unobservedSet}
                 {\bfS\tn^\unobservedSet}$ as follows:
      \FOR{$\jq = 1, \cdots, \nQ$}
      \IF{$\xonq^\unobservedSet$ is observed}
          \STATE $\mu^\unobservedSet\inq{\jn}{\jq}=\xonq^\unobservedSet$
            and  $(\bfS\tn^\unobservedSet)\inq{\jq}{\jq} = \varepsilon, \text{where: } \varepsilon \rightarrow 0$ 
          \STATE Fix $\mu^\unobservedSet\inq{\jn}{\jq}, (\bfS\tn^\unobservedSet)\inq{\jq}{\jq}$ in the optimiser
           \COMMENT{\emph{(\ie they don't constitute parameters)}}
      \ELSE
            \STATE $\mu_{\jn, \jq}^\unobservedSet = \hat{\mu}_{\jn, \jq}^\unobservedSet$
              and  $(\bfS\tn^\unobservedSet)\inq{\jq}{\jq} = (\hat{\bfS}\tn^\unobservedSet)\inq{\jq}{\jq}$
      \ENDIF
      \ENDFOR
\ENDFOR \label{state:forEnds}

\SSTATE Train model $\mathcal{M}^{\observedSet, \unobservedSet}$ with inputs $q(\mX^{\{\observedSet,\unobservedSet\}} | \mXo^{\{\observedSet,\unobservedSet\}})$ and outputs $(\mY^\observedSet, \mY^\unobservedSet)$. The input distribution $q(\mX^{\{\observedSet,\unobservedSet\}} | \mXo^{\{\observedSet,\unobservedSet\}}) = q(\mX^\observedSet|\mXo^\observedSet)q(\mX^\unobservedSet|\mXo^\unobservedSet)$ is constructed in steps \ref{state:qObs}, \ref{state:forStart}-\ref{state:forEnds} and further optimised in the non-fixed locations.
\SSTATE Model $\mathcal{M}^{\observedSet, \unobservedSet}$ now constitutes the semi-described GP and can be used for all required prediction tasks.
\end{algorithmic}
\end{algorithm*}

Algorithm \ref{algorithm:semiSupervised} outlines the approach in more detail.
This formulation defines a \emph{semi-described GP} approach which naturally incorporates fully and partially observed examples by communicating the uncertainty throughout the relevant parts of the model in a principled way. 
Indeed, the predictive uncertainty obtained when imputing missing values in the first step of the pipeline is incorporated as input uncertainty in the second step of the pipeline. In extreme cases resulting in very non-confident predictions, for example in the presence of outliers, the corresponding locations will simply be ignored automatically due to the large uncertainty. This mechanism, together with the subsequent optimisation of the parameters of $q(\mX^\unobservedSet | \mXo^\unobservedSet)$ in stage 2, guards against reinforcing bad predictions when imputing missing values based on a smaller training set.  The model includes GP regression and the GP-LVM as special cases. In particular, in the limit of having no observed values our semi-described GP is equivalent to the GP-LVM and when there are no missing values it is equivalent to GP regression.

There are some similarities to traditional self-training \citep{Rosenberg:semiSupervisedSelfTraining}, but as there are no straightforward mechanisms to propagate uncertainty in that domain, they typically rely on boot-strapping followed by thresholding ``bad'' samples to prevent model over-confidence.
In our framework, the predictions made by the initial model only constitute initialisations which are later optimised along with model parameters and, hence, we refer to this step as \emph{partial} self-training. Further, the predictive uncertainty is not used as a hard measure of discarding unconfident predictions; instead, we allow all values to contribute according to an optimised uncertainty measure, that is, the input variances $\bfS\tn$. Therefore, the way in which uncertainty is handled makes the self-training part of our algorithm principled compared to many bootstrap-based approaches. 

\subsubsection*{DEMONSTRATION}

\begin{figure*}[t]
\begin{center}
       \includegraphics[width=0.98\textwidth]{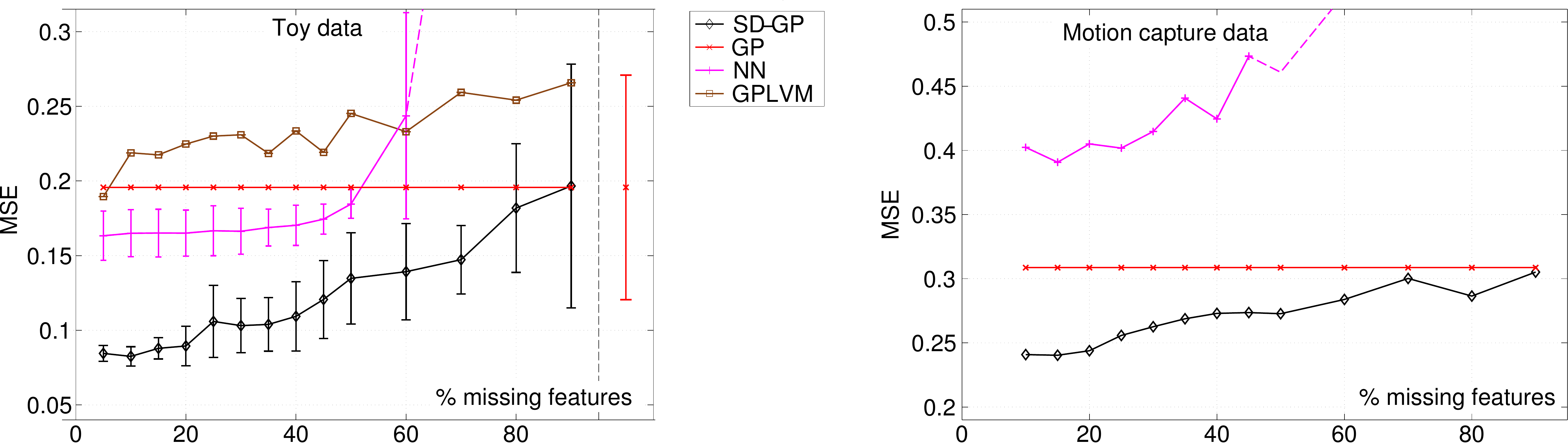}
\end{center}
\vspace{-2pt}
\caption[Semi-described GP: partially missing inputs experiment.]{
MSE for predictions obtained by different methods on semi-described learning. GP cannot handle partial observations, thus the uncertainty ($2\sigma$) is constant; for clarity, the errorbar is plotted separately on the right of the dashed vertical line (for nonsensical $x$ values). The results for simulated data are obtained from 4 trials. For clarity, the limits on the $y-$axis are fixed, so when the errors become too big for certain methods they get off the chart. The errorbars for the GPLVM-based approach are also too large and not plotted. The full picture is given in figure \ref{fig:semiSupervisedSuppl} (Appendix).}
\label{fig:semiSupervised}
\vspace{-2pt}
\end{figure*}

\noindent We considered simulated and real-world data to demonstrate our semi-described GP algorithm. The simulated data were created by sampling inputs $\mXo$ from a GP (which was unknown to the competing models) and then giving these samples as input to another unknown GP, to obtain corresponding outputs $\mY$. For the real-world data demonstration we considered a motion capture dataset taken from subject 35 in the CMU motion capture database. We selected a subset of walk and run motions of a human body represented as a set of $59$ joint locations. We formulated a regression problem where the first $20$ dimensions of the original data are used as targets and the remaining $39$ as inputs. That is, given a partial joint representation of the human body, the task is to infer the rest of the representation. For both datasets, simulated and motion capture, we selected a portion of the training inputs, denoted as $\mXo^\unobservedSet$, to have randomly missing features. 
The extended dataset $\left( \left(\mXo^\observedSet, \mXo^\unobservedSet \right),\left(\mY^\observedSet, \mY^\unobservedSet \right) \right)$ was used to train: a) our method, referred to as semi-described GP (SD-GP) b) multiple linear regression (MLR) c) regression by performing nearest neighbour (NN) search between the test and training instances, in the observed input locations d) performing data imputation using the standard GP-LVM. Not taking into account the predictive uncertainty during imputation was found to have catastrophic results in the simulated data, as the training set was not robust against bad predictions. Therefore, the ``GP-LVM'' variant was not considered in the real data experiment. We also considered: e) a standard GP  which cannot handle missing inputs straightforwardly and so was trained only on the observed data $\left(\mXo^\observedSet, \mY^\observedSet \right)$. 
The goal was to reconstruct test outputs $\mY_*$ given fully observed test inputs $\mXo_*$. For the simulated data we used the following sizes: $| \mXo^\observedSet | = 40$, $| \mXo^\unobservedSet | = 60$ and $| \mXo_* | = 100 $. The dimensionality of the inputs is $\nQ = 15$ and of the outputs is $\nD = 5$. For the motion capture data we used $| \mXo^\observedSet | = 50$, $| \mXo^\unobservedSet | = 80$ and $| \mXo_* | = 200 $. In fig. \ref{fig:semiSupervised} we plot the MSE obtained by the competing methods for a varying percentage offv missing features in $\mXo^\unobservedSet$. For the simulated data experiment, each of the points in the plot is an average of 4 runs which considered different random seeds.
For clarity, the $y-$axis limit is fixed in figure \ref{fig:semiSupervised}, because some methods produced huge errors. The full picture is in figure \ref{fig:semiSupervisedSuppl} (Appendix).
As can be seen in the figures, the semi-described GP is able to handle the extra data and make much better predictions, even if a very large portion is missing. Indeed, its performance starts to converge to that of a standard GP when there are 90\% missing values in $\mXo^\unobservedSet$ and performs identically to the standard GP when 100\% of the values are missing. We found that when $\nQ$ is large compared to $\nD$ and $\nN$, then the data imputation step can be problematic as the percentage of missing features in $\mXo^\unobservedSet$ approaches $100\%$ \ie the method is reliant on having some covariates available. Appendix \ref{app:effectQDN} discusses this behaviour, but a more systematic investigation is left as future work.

\subsection{AUTO-REGRESSIVE GAUSSIAN PROCESSES \label{subsec:autoregressive}}

Having a method which implicitly models the uncertainty in the inputs of a GP also allows for doing predictions in an autoregressive manner \citep{Oakley:computer02} while propagating the uncertainty through the predictive sequence \citep{Girard:uncertain01Compact,quinonero2003propagation}. Specifically, assuming that the given data $\mY$ constitute a multivariate timeseries where the observed time vector $\bft$ is equally spaced, and given a time-window of length $\tau$, we can reformat $\mY$ into input-output collections of pairs $\hat{\mXo}$ and $\hat{\mY}$ as follows:
the first input to the model, $\hat{\vxo}\inn{1}$, will be given by the stacked vector $\left[\vy\inn{1}, ...,  \vy\inn{\tau} \right]$ and the first output, $\hat{\vy}\inn{1}$, will be given by $\vy\inn{\tau+1}$ and similarly for the other data in $\hat{\mXo}$ and $\hat{\mY}$, so that:
\begin{align*}
[\hat{\vxo}\inn{1}, \hat{\vxo}\inn{2}, ..., \hat{\vxo}\inn{\nN-\tau}] &= \\
\big[ [\vy\inn{1}, \vy\inn{2}, ..., & \vy\inn{\tau} ],  \left[\vy\inn{2}, \vy\inn{3}, ..., \vy\inn{\tau+1}\right], ... \big], \\ 
[\hat{\vy}\inn{1}, \hat{\vy}\inn{2}, ..., \hat{\vy}\inn{\nN-\tau}] &= [\vy\inn{\tau+1}, \vy\inn{\tau+2}, ..., \vy\inn{\nN}]. 
\end{align*}

To perform extrapolation we first train the model on the modified dataset $(\hat{\mXo}, \hat{\mY})$. By referring to the semi-described formulation described in Section \ref{subsec:semiDescribed}, we assign all training inputs to the observed set $\observedSet$. After training, we can perform iterative prediction to find a future sequence $\hat{\mXo}_* \coloneqq \left[\vy\inn{\nN+1}, \vy\inn{\nN+2},  ... \right]$ where, similarly to the approach taken by \cite{Girard:uncertain01Compact}, the predictive variance in each step is accounted for and propagated in the subsequent predictions.  The algorithm makes iterative 1-step predictions in the future; initially, the output $\hat{\vxo}\innTs{1} \coloneqq \vy\inn{\nN+1}$ will be predicted (given the training set) with predictive variance $\hat{\bfS}_{*;1}$. In the next step, the ``observations'' set will be augmented to include the distribution of predictions over $\vy\inn{\nN+1}$, by defining $q(\vx\inn{\nN + 1} | \hat{\vxo}\innTs{1}) = \gaussianDist{\vx\inn{\nN + 1}}{\hat{\vxo}_{*,1}}{\hat{\bfS}_{*;1}}$, and so on. This simulation process can be seen as constructing a predictive sequence step by step, \ie the newly inserted input points constitute parts of the (test) predictive sequence and not training points. Therefore, this procedure can be seen as an iterative version of semi-described learning.

Note that it is straightforward to extend this model by applying this auto-regressive mechanism in a latent space of a stacked model or, more generally, as a deep GP \citep{Damianou:deepGPs13Compact}. By additionally introducing functions that map from this latent space nonlinearly to an observation space, we obtain a fully nonlinear state space model in the manner of \cite{deisenroth2012robust}. For our model, uncertainty is encoded in both the states and the nonlinear transition functions. Correct propagation of uncertainty is vital in well calibrated models of future system behavior, and automatic determination of the structure of the model (\eg the window size) can be informative in describing the order of the underlying dynamical system.
\vspace{-1pt}

\subsubsection*{DEMONSTRATION: ITERATIVE FORECASTING}

\begin{figure*}[t]
\begin{center}
       \includegraphics[width=1\textwidth]{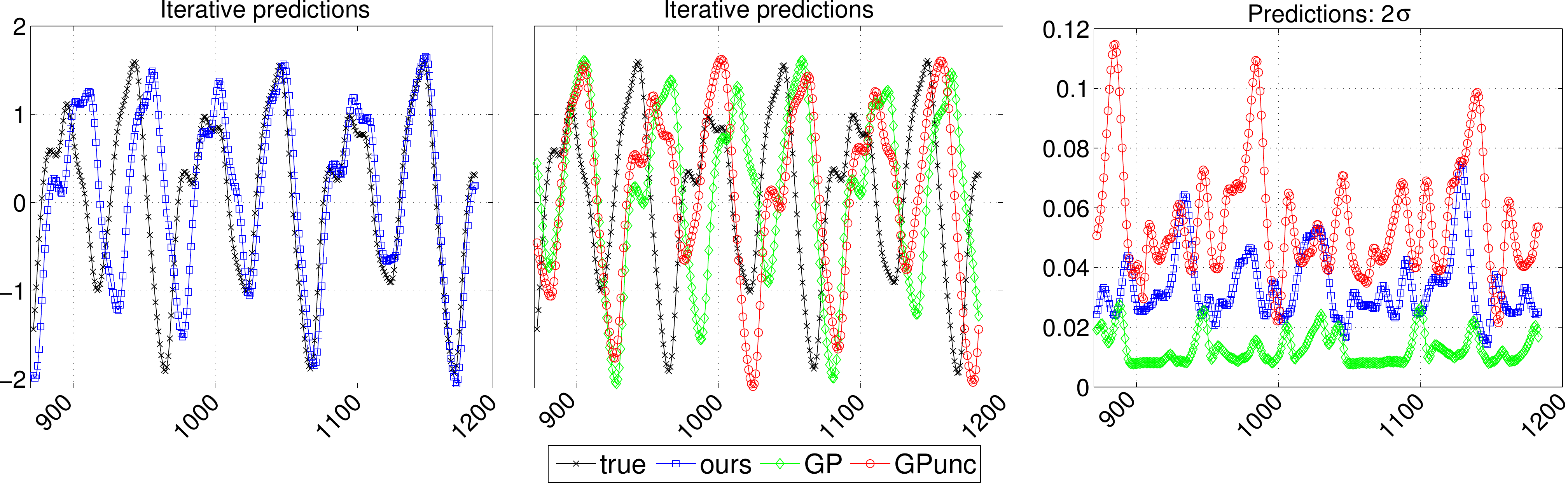}
\end{center}
\vspace{-3pt}
\caption[Auto-regressive GPs: Iterative prediction for a chaotic timeseries.]{
Chaotic timeseries: forecasting $1110$ steps ahead by iterative prediction. The first $800$ steps are not shown here, but figure \ref{fig:uncInputsExtrapolationMore} (Appendix) gives the complete picture.
Comparing: a ``naive autoregressive'' GP which does not propagate (and hence underestimates) the uncertainties; the method of \citet{Girard:uncertain01Compact}, referred to as $\text{GP}_{\text{uncert}}$; and our approach, which closely tracks the true test sequence until the last steps of the extrapolation. The comparative depiction of the predictions is split into two plots (for clarity), left and center. The rightmost plot shows the predictive uncertainties ($2 \sigma$). 
$x-$axis is the prediction step ($t$) and $y-$axis is the function value, $f(t)$.
}
\vspace{-5pt}
\label{fig:uncInputsExtrapolation}
\end{figure*}

Here we demonstrate our framework in the simulation of a state space model. We consider the Mackey-Glass chaotic time series, a standard benchmark which was also considered by \cite{Girard:uncertain01Compact}. The data is one-dimensional so that the timeseries can be represented as pairs of values $\{\vy, \bft\}, t = 1,2, \cdots, \nN$ and simulates the process:
\begin{align*}
\frac{\intd \zeta(t)}{dt} = 
-b\zeta(t) + \alpha \mathsmaller{\frac{\zeta(t-T)}{1+\zeta(t-T)^{10}}}, (\alpha,b,T) = (0.2,0.1,17).
\end{align*}

Obviously the generating process is very non-linear, rendering this dataset challenging. We trained the autoregressive model on data from this series, where the modified dataset $\{ \hat{\vxo}, \hat{\vy} \}$ was created with $\tau = 18$ and we used the first $4\tau=72$ points to train the model and predicted the subsequent $1110$ points through iterative free simulation. 

We compared our method with a ``naive autoregressive'' GP model where the input-output pairs were given by the autoregressive modification of the dataset $\{\hat{\vxo}, \hat{\vy}\}$. For that model, the predictions are made iteratively and the predicted values after each predictive step are added to the ``observation'' set. However, this standard GP model has no straight forward way of incorporating/propagating the uncertainty and, therefore, the input uncertainty is zero for every step of the iterative predictions.
We also compared against the method of \cite{Girard:uncertain01Compact}\footnote{We implemented the basic moment matching approach, although in the original paper the authors use additional approximations, namely Taylor expansion around the predictive moments.}, denoted in the plots as ``$\text{GP}_{\text{uncert}}$''.
Figure \ref{fig:uncInputsExtrapolation} shows the results for the last 310 steps (\ie $t=800$ onwards) of the full free simulation ($1110-$step ahead forecasting); figure \ref{fig:uncInputsExtrapolationMore} (Appendix) gives a more complete picture.
As can be seen in the variances plot, both our method and $\text{GP}_{\text{uncert}}$ are more robust in handling the uncertainty throughout the predictions; the ``naive'' GP method underestimates the uncertainty. Consequently, as can be seen in figure \ref{fig:uncInputsExtrapolationMore}, in the first few predictions all methods give the same answer. However, once the predictions of the ``naive'' method diverge a little by the true values, the error is carried on and amplified due to underestimating the uncertainty. On the other hand, $\text{GP}_{\text{uncert}}$ perhaps overestimates the uncertainty and, therefore, is more conservative in its predictions, resulting in higher errors. Quantification of the error is shown in Table \ref{table:uncInputsError} (Appendix).

\subsection{SEMI-SUPERVISED LEARNING\label{subsec:partialOutputs}}

\begin{figure*}[ht]
\begin{center}
       \includegraphics[width=0.79\textwidth]{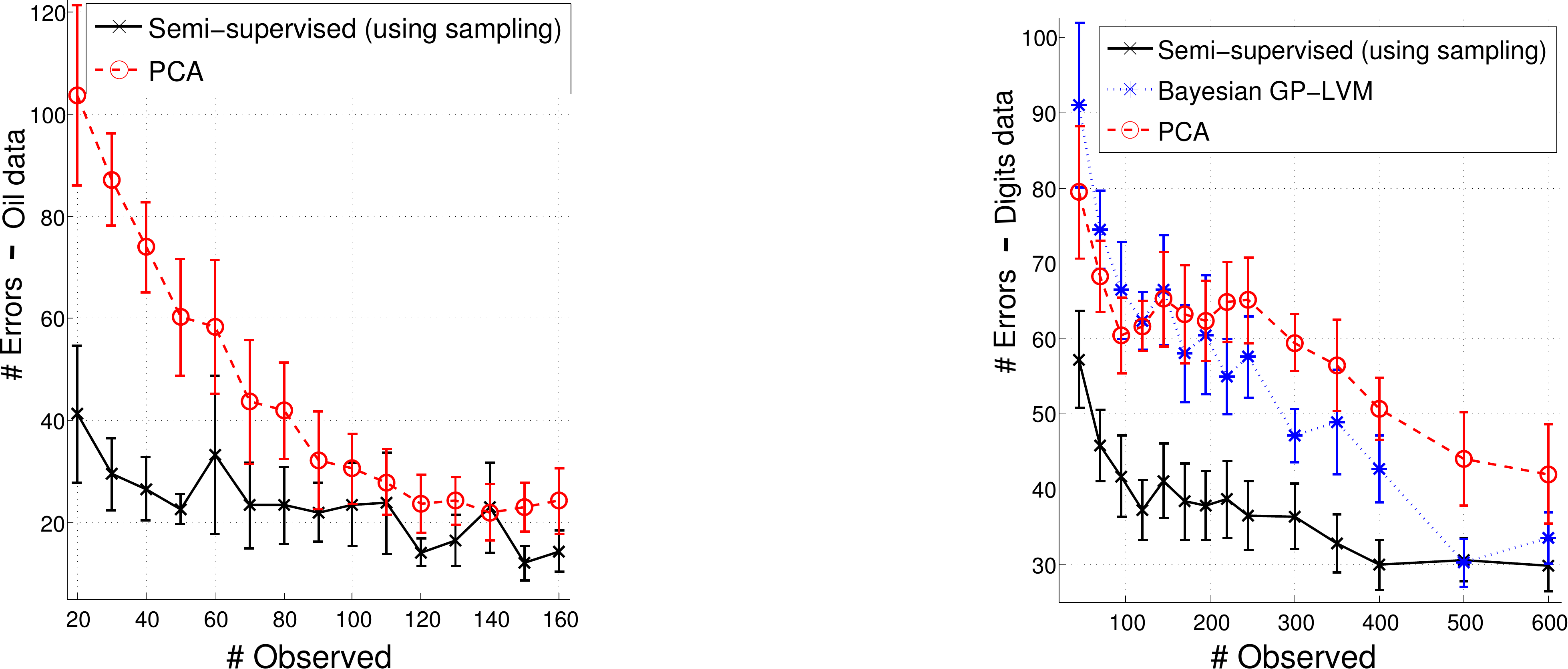}
       \label{subfig:digits}
\end{center}
\vspace{-4pt}
\caption{
Plots of the number of incorrectly classified test points as a function of $| \bfZ^\labelledSet |$. Multiple trials were performed, but the resulting errorbars are shown at one standard deviation. In small training sets large errorbars are expected because, occasionally, very challenging instances/outliers can be included and result in high error rates (for all methods) that affect the overall standard deviation. The Bayesian GP-LVM baseline struggled with small training sets and performed very badly in the oil dataset; thus, it is not plotted for clarity.}
\vspace{-3pt}
\label{fig:missingOutputs}
\end{figure*}

In this section we study semi-supervised learning which, in contrast to semi-described learning, is for handling missing values in the outputs. This scenario is typically encountered in classification settings. 
Therefore, we introduce the sets $\{ \labelledSet, \unlabelledSet \}$ that index respectively the \emph{labelled} and \emph{missing} (unlabelled) rows of the outputs (labels) $\mY$. Accordingly, the full dataset is split so that $\mXo = (\mXo^\labelledSet, \mXo^\unlabelledSet)$ and $\mY = (\mY^\labelledSet, \mY^\unlabelledSet)$, where $\mXo$ is now fully observed.
The task is then to devise a method that improves classification performance by using both labelled and unlabelled data.

Inspired by \cite{Kingma14:semiSupervisedDeepGenerative} we define a semi-supervised GP framework where features are extracted from all available information and, subsequently, are given as inputs to a discriminative classifier. Specifically, using the whole input space $\mXo$, we learn a low-dimensional latent space $\mX$ through an approximate posterior $q(\mX) \approx p(\mX | \mXo)$. Obviously, this specific case where the input space is uncertain but totally unobserved (\ie a latent space) just reduces to the Bayesian GP-LVM model. Notice that the posterior $q(\mX)$ is no longer constrained with $\mXo$ but, rather, directly approximates $p(\mX | \mXo)$, since we now have a forward \emph{probabilistic} mapping from $\mX$ to $\mXo$ and $\mXo$ is treated as a random variable with $p(\mXo | \mX)$ being a Gaussian distribution, \ie exactly the same setting used in the GP-LVM.
Since there is one-to-one correspondence between $\mX$, $\mXo$ and $\mY$, we can notationally write $\mX = (\mX^\labelledSet, \mX^\unlabelledSet)$. Further, since we assume that $q(\mX)$ is factorised across datapoints, we can write $q(\mX) = q(\mX^\labelledSet)q(\mX^\unlabelledSet)$.

In the second step of our semi-supervised algorithm, we train a discriminative classifier from $q(\mX^\labelledSet)$ to the observed labelled space, $\mY^\labelledSet$. The main idea is that, by including the inputs $\mXo^\unlabelledSet$ in the first learning step, we manage to define a better latent embedding from which we can extract a more useful set of features for the discriminative classifier. Notice that what we would ideally use as input to the discriminative classifier is a whole distribution, rather than single point estimates. Therefore, we wish to take advantage of the associated uncertainty; specifically, we can populate the labelled set by sampling from the distribution $q(\mX^\labelledSet)$. For example, if a latent point $\xn^\labelledSet$ corresponds to the input-output pair $(\xon^\labelledSet, \yn^\labelledSet)$, then a sample from $q(\xn^\labelledSet)$ will be assigned the label $\yn^\labelledSet$. 

The two inference steps described above are graphically depicted in Figure \ref{subfig:model3}.
This is exactly the same setting suggested by  \cite{Kingma14:semiSupervisedDeepGenerative}, but here we wish to investigate its applicability in a non-parametric, Gaussian process based framework. The very encouraging results reported below point towards the future direction of applying this technique in the framework of deep Gaussian processes \citep{Damianou:deepGPs13Compact}, so as to be able to compare to \citep{Kingma14:semiSupervisedDeepGenerative} who considered deep, generative (but nevertheless parametric) models.

\subsubsection*{DEMONSTRATION}

We evaluated our semi-supervised GP algorithm in two datasets: firstly, we considered 2000 examples from the
USPS handwritten digit database \citep{Hull:CEDAR94}. These examples contained the digits $\{ 0,2,4,6 \}$ and were split so that 800 instances were used as a test set. From the remaining 1200 instances, we selected various portions to be labelled and the rest to be unlabelled. The experiment was repeated 8 times (each time involving different subsets due to different random seeds), so that we can include errorbars in our plots. Secondly, we considered the oil flow data \citep{Bishop:oil93Compact} that consist of 1000, 12 dimensional observations belonging to three known classes corresponding to different phases of oil flow. In each of the 10 performed trials, 700 instances were used as a test set whereas the rest were split to different proportions of labelled/unlabelled sets. Multi-label data can also be handled by our method, but this case was not considered here.

Our method learned a low-dimensional embedding $q(\mX)$ from all available inputs, and a logistic regression classifier was then trained from the relevant parts of the embedding to the corresponding class space. We experimented with taking different numbers of samples from $q(\mX^\labelledSet)$ for populating the initial labelled set; the difference after increasing over 6 samples was minimal. Also, when using only the mean of $q(\mX^\labelledSet)$ (as opposed to using multiple samples) we obtained worse results (especially in the digits data), but this method still outperformed the baselines.
We compared with training the classifier on features learned by (a) a standard Bayesian GP-LVM and (b) PCA, both applied in $\mXo^\labelledSet$. Both of the baselines do not take $\mXo^\unlabelledSet$ into account, nor do they populate small training sets using sampling. Figure \ref{fig:missingOutputs} presents results suggesting that our approach manages to effectively take into account unlabelled data. The gain in performance is significant, and our method copes very well even when labelled data is extremely scarce. Notice that all methods would perform better if a more robust classifier was used, but logistic regression was a convenient choice for performing multiple trials fast. Therefore, our conclusions can be safely drawn from the obtained relative errors, since all methods were compared on equal footing.

\section{DISCUSSION AND FUTURE WORK \label{sec:conclusion}}
We have defined semi-described learning as the scenario where missing and uncertain values occur in the inputs. We considered semi-described problems to be part of a general class of missing value problems that also includes semi-supervised learning and auto-regressive future state simulation. 
A principled method for including input uncertainty and partial inputs in Gaussian process models was also introduced to solve these problems within a single, coherent framework. We explicitly represent this uncertainty as approximate posterior distributions which are variationally constrained. This allowed us to further define algorithms for casting the missing value problems as particular instances of learning pipelines which use our variationally constrained GP formulation as a building block. Our algorithms resulted in significant performance improvement in forecasting, regression and classification.
We believe that our contribution paves the way for building powerful models for representation learning from real-world, heterogenous data. In particular, this can be achieved by combining our method with deep Gaussian process models \citep{Damianou:deepGPs13Compact} that use relevance determination techniques \citep{Damianou:mrd12Compact}, so as to consolidate semi-described hierarchies of features that are gradually abstracted to concepts. 
We plan to investigate the application of
these models in settings where control
\citep{Deisenroth:dataEfficient14Compact} or robotic systems learn 
by simulating future states in an auto-regressive manner and by
using incomplete data with miminal human intervention. Transfer learning is another promising direction for applying these models.

\subsubsection*{ACKNOWLEDGEMENTS}
This research was funded by the European research project EU FP7-ICT (Project Ref
612139 ``WYSIWYD''). We thank Michalis Titsias for useful discussions.

\bibliographystyle{abbrvnat}
\bibliography{references,other,lawrence,zbooks} 

\begin{thebibliography}{29}
\providecommand{\natexlab}[1]{#1}
\providecommand{\url}[1]{\texttt{#1}}
\expandafter\ifx\csname urlstyle\endcsname\relax
  \providecommand{\doi}[1]{doi: #1}\else
  \providecommand{\doi}{doi: \begingroup \urlstyle{rm}\Url}\fi

\bibitem[Bishop and James(1993)]{Bishop:oil93Compact}
C.~M. Bishop and G.~D. James.
\newblock Analysis of multiphase flows using dual-energy gamma densitometry and
  neural networks.
\newblock \emph{Nuclear Instruments and Methods in Physics Research},
  A327:\penalty0 580--593, 1993.

\bibitem[Chapelle et~al.(2006)Chapelle, Sch\"olkopf, and
  Zien]{Chapelle:semisuper06}
O.~Chapelle, B.~Sch\"olkopf, and A.~Zien, editors.
\newblock \emph{Semi-supervised Learning}.
\newblock MIT Press, Cambridge, MA, 2006.

\bibitem[Dai et~al.(2014)Dai, Damianou, Hensman, and Lawrence]{Dai2014Parallel}
Z.~Dai, A.~Damianou, J.~Hensman, and N.~Lawrence.
\newblock Gaussian process models with parallelization and {GPU} acceleration.
\newblock \emph{arXiv preprint arXiv:1410.4984}, 2014.

\bibitem[Dallaire et~al.(2009)Dallaire, Besse, and
  Chaib-Draa]{dallaire2009learning}
P.~Dallaire, C.~Besse, and B.~Chaib-Draa.
\newblock Learning {G}aussian process models from uncertain data.
\newblock In \emph{Neural Information Processing}, pages 433--440. Springer,
  2009.

\bibitem[Damianou and Lawrence(2013)]{Damianou:deepGPs13Compact}
A.~Damianou and N.~Lawrence.
\newblock Deep {G}aussian processes.
\newblock In \emph{Proceedings of the Sixteenth International Workshop on
  Artificial Intelligence and Statistics (AISTATS)}, pages 207--215. JMLR W\&CP
  31, 2013.

\bibitem[Damianou and Lawrence(2014)]{Damianou:pipelines14}
A.~Damianou and N.~Lawrence.
\newblock Uncertainty propagation in {G}aussian process pipelines.
\newblock \emph{NIPS workshop on modern non-parametrics}, 2014.

\bibitem[Damianou et~al.(2011)Damianou, Titsias, and
  Lawrence]{Damianou:vgpds11Compact}
A.~Damianou, M.~Titsias, and N.~D. Lawrence.
\newblock Variational {G}aussian process dynamical systems.
\newblock In \emph{Advances in Neural Information Processing Systems 24}, pages
  2510--2518. 2011.

\bibitem[Damianou et~al.(2012)Damianou, Ek, Titsias, and
  Lawrence]{Damianou:mrd12Compact}
A.~Damianou, C.~Ek, M.~Titsias, and N.~Lawrence.
\newblock Manifold relevance determination.
\newblock In \emph{Proceedings of the 29th International Conference on Machine
  Learning (ICML)}, pages 145--152. Omnipress, 2012.

\bibitem[Deisenroth et~al.(2012)Deisenroth, Turner, Huber, Hanebeck, and
  Rasmussen]{deisenroth2012robust}
M.~P. Deisenroth, R.~D. Turner, M.~F. Huber, U.~D. Hanebeck, and C.~E.
  Rasmussen.
\newblock Robust filtering and smoothing with {G}aussian processes.
\newblock \emph{Automatic Control, IEEE Transactions on}, 57\penalty0
  (7):\penalty0 1865--1871, 2012.

\bibitem[Deisenroth et~al.(2014)Deisenroth, Fox, and
  Rasmussen]{Deisenroth:dataEfficient14Compact}
M.~P. Deisenroth, D.~Fox, and C.~E. Rasmussen.
\newblock Gaussian processes for data-efficient learning in robotics and
  control.
\newblock \emph{IEEE Transactions on Pattern Analysis and Machine
  Intelligence}, 99:\penalty0 1, 2014.
\newblock ISSN 0162-8828.

\bibitem[Ek et~al.(2008)Ek, Rihan, Torr, Rogez, and Lawrence]{Ek:ambiguity08}
C.~H. Ek, J.~Rihan, P.~Torr, G.~Rogez, and N.~D. Lawrence.
\newblock Ambiguity modeling in latent spaces.
\newblock In A.~{Popescu-Belis} and R.~Stiefelhagen, editors, \emph{Machine
  Learning for Multimodal Interaction (MLMI 2008)}, LNCS, pages 62--73.
  Springer-Verlag, 28--30 June 2008.

\bibitem[Gal et~al.(2014)Gal, van~der Wilk, and Rasmussen]{Gal:Distributed14}
Y.~Gal, M.~van~der Wilk, and C.~E. Rasmussen.
\newblock Distributed variational inference in sparse {G}aussian process
  regression and latent variable models.
\newblock \emph{arXiv:1402.1389}, 2014.

\bibitem[Ghahramani and Jordan(1994)]{Ghahramani:missing94}
Z.~Ghahramani and M.~I. Jordan.
\newblock Learning from incomplete data.
\newblock Technical Report CBCL 108, Massachusetts Institute of Technology,
  1994.

\bibitem[Girard et~al.(2003)Girard, Rasmussen, {Qui\~nonero Candela}, and
  Murray-Smith]{Girard:uncertain01Compact}
A.~Girard, C.~E. Rasmussen, J.~{Qui\~nonero Candela}, and R.~Murray-Smith.
\newblock Gaussian process priors with uncertain inputs---application to
  multiple-step ahead time series forecasting.
\newblock In \emph{Advances in Neural Information Processing Systems}, pages
  529--536, 2003.

\bibitem[Hull(1994)]{Hull:CEDAR94}
J.~J. Hull.
\newblock A database for handwritten text recognition research.
\newblock \emph{IEEE Transactions on Pattern Analysis and Machine
  Intelligence}, 16:\penalty0 550--554, 1994.

\bibitem[Kingma et~al.(2014)Kingma, Rezende, Mohamed, and
  Welling]{Kingma14:semiSupervisedDeepGenerative}
D.~P. Kingma, D.~J. Rezende, S.~Mohamed, and M.~Welling.
\newblock Semi-supervised learning with deep generative models.
\newblock \emph{CoRR}, abs/1406.5298, 2014.

\bibitem[Lawrence(2006)]{Lawrence:gplvmtut06}
N.~D. Lawrence.
\newblock The {G}aussian process latent variable model.
\newblock Technical Report CS-06-03, The University of Sheffield, Department of
  Computer Science, 2006.

\bibitem[Lawrence and Jordan(2005)]{Lawrence:semisuper04}
N.~D. Lawrence and M.~I. Jordan.
\newblock Semi-supervised learning via {G}aussian processes.
\newblock In L.~Saul, Y.~Weiss, and L.~Bouttou, editors, \emph{Advances in
  Neural Information Processing Systems}, volume~17, pages 753--760, Cambridge,
  MA, 2005. MIT Press.

\bibitem[Lawrence and {Qui\~nonero Candela}(2006)]{Lawrence:backconstraints06}
N.~D. Lawrence and J.~{Qui\~nonero Candela}.
\newblock Local distance preservation in the {GP-LVM} through back constraints.
\newblock In W.~Cohen and A.~Moore, editors, \emph{Proceedings of the
  International Conference in Machine Learning}, volume~23, pages 513--520.
  Omnipress, 2006.
\newblock ISBN 1-59593-383-2.
\newblock \doi{10.1145/1143844.1143909}.

\bibitem[McHutchon and Rasmussen(2011)]{mchutchon:gaussian}
A.~McHutchon and C.~E. Rasmussen.
\newblock Gaussian process training with input noise.
\newblock In \emph{NIPS}, 2011.

\bibitem[Oakley and O'Hagan(2002)]{Oakley:computer02}
J.~Oakley and A.~O'Hagan.
\newblock Bayesian inference for the uncertainty distribution of computer model
  outputs.
\newblock \emph{Biometrika}, 89\penalty0 (4):\penalty0 769--784, 2002.

\bibitem[Osborne and Roberts(2007)]{Osborne:gpreport2007}
M.~Osborne and S.~J. Roberts.
\newblock Gaussian processes for prediction.
\newblock Technical report, Department of Engineering Science, University of
  Oxford, 2007.

\bibitem[Qui{\~n}onero-Candela(2004)]{quinonero2004learning}
J.~Qui{\~n}onero-Candela.
\newblock \emph{Learning with uncertainty-Gaussian processes and relevance
  vector machines}.
\newblock PhD thesis, Technical University of Denmark, 2004.

\bibitem[Quinonero-Ca{\~n}dela and Roweis(2003)]{Candela:clearning03}
J.~Quinonero-Ca{\~n}dela and S.~Roweis.
\newblock Data imputation and robust training with {G}aussian processes.
\newblock \emph{NIPS}, 2003.

\bibitem[Qui{\~n}onero-Candela et~al.(2003)Qui{\~n}onero-Candela, Girard,
  Larsen, and Rasmussen]{quinonero2003propagation}
J.~Qui{\~n}onero-Candela, A.~Girard, J.~Larsen, and C.~E. Rasmussen.
\newblock Propagation of uncertainty in bayesian kernel models-application to
  multiple-step ahead forecasting.
\newblock In \emph{Acoustics, Speech, and Signal Processing, 2003.
  Proceedings.(ICASSP'03). 2003 IEEE International Conference on}, volume~2,
  pages II--701. IEEE, 2003.

\bibitem[Rosenberg et~al.(2005)Rosenberg, Hebert, and
  Schneiderman]{Rosenberg:semiSupervisedSelfTraining}
C.~Rosenberg, M.~Hebert, and H.~Schneiderman.
\newblock Semi-supervised self-training of object detection models.
\newblock In \emph{Application of Computer Vision, 2005. WACV/MOTIONS '05
  Volume 1.}, volume~1, pages 29--36, Jan 2005.
\newblock \doi{10.1109/ACVMOT.2005.107}.

\bibitem[Rubin(2004)]{rubin2004multiple}
D.~B. Rubin.
\newblock \emph{Multiple imputation for nonresponse in surveys}, volume~81.
\newblock John Wiley \& Sons, 2004.

\bibitem[Titsias and Lawrence(2010)]{BayesianGPLVM}
M.~Titsias and N.~D. Lawrence.
\newblock Bayesian {G}aussian process latent variable model.
\newblock \emph{Journal of Machine Learning Research - Proceedings Track},
  9:\penalty0 844--851, 2010.

\bibitem[Zhu et~al.(2012)Zhu, Ahmed, and Xing]{zhu2012medlda}
J.~Zhu, A.~Ahmed, and E.~P. Xing.
\newblock Medlda: maximum margin supervised topic models.
\newblock \emph{The Journal of Machine Learning Research}, 13\penalty0
  (1):\penalty0 2237--2278, 2012.

\end{thebibliography}

\clearpage
\appendix
\section{APPENDIX: VARIATIONAL LOWER BOUND}

In this appendix we give some more details on the computation of the variational lower bound for the variationally constrained model.

The augmented joint probability density (after introducing the inducing points) takes the form,
\begin{align*}
& p(\dataMatrix,\mappingFunctionMatrix, \inducingMatrix,\latentMatrix | \bfX_u) \nonumber \\  
& \qquad = p(\dataMatrix|\mappingFunctionMatrix) 
   p(\mappingFunctionMatrix|\inducingMatrix,\latentMatrix,\latentMatrix_u)
   p(\inducingMatrix|\latentMatrix_u) p(\latentMatrix) \nonumber \\
& \qquad =  \left( \prod_{\outputIndex=1}^\dataDim 
p(\dataVector_\outputIndex | \mappingFunctionVector_\outputIndex) p(\mappingFunctionVector_\outputIndex | \inducingVector_\outputIndex, \latentMatrix, \latentMatrix_u) 
p(\inducingVector_\outputIndex | \latentMatrix_u) \right) p(\latentMatrix) .  
\end{align*}
In the r.h.s above, the observed inputs $\bfZ$ do not appear, exactly because we introduce them through the variational constraint, which does not constitute a probabilistic mapping. In the above equations we have
\begin{align*}
p(\mappingFunctionVector_\outputIndex | \inducingVector_\outputIndex,\latentMatrix,\latentMatrix_u) = \gaussianDist{\mappingFunctionVector_\outputIndex}{\mathbf{a}_\outputIndex}{\bfSigma_f} \label{priorF2},
\end{align*}
being the conditional GP prior with
\begin{equation*}
\label{eq:conditionalGPmeanCovar}
\mathbf{a}_j = \Kfu \Kuu^{-1} \inducingVector_\outputIndex \text{\; \; and \; \; }
\bfSigma_f = \Kff - \Kfu \Kuu^{-1} \Kuf
\end{equation*}
and
\begin{equation*}
\label{pfu}
p(\inducingVector_\outputIndex|\latentMatrix_u) = \mathcal{N}(\inducingVector_\outputIndex|\mathbf{0},
\Kuu),
\end{equation*}
is the marginal GP
prior over the inducing variables. In the above expressions, $\Kuu$ denotes
the covariance matrix constructed by evaluating the covariance function 
on the inducing points, $\Kuf$ is the cross-covariance between the inducing
and the latent points and $\Kfu = \Kuf^\top$.

In order to perform variational inference in this expanded probability model, we introduce 
the variational distributions $q(\bfX|\bfZ)$ and $q(\bfU)$, which are both taken to be Gaussian.
For convenience, we drop the inducing points $\bfX_u$ from our expressions for the remainder of the Appendix, for convenience.
We now have:
\begin{align*}
& \log p(\bfY | \bfX_u) = \nonumber \\ 
& \qquad \log \int_{\bfU, \bfX} p(\bfU) p(\bfX) \int_\bfF p(\bfY | \bfF) p(\bfF|\bfU, \bfX) .
\end{align*}

By applying Jensen's inequality, we obtain a lower bound $\F(q(\bfX),q(\bfU))$ on the above marginal likelihood, where:
\begin{align}
&\F(q(\bfX|\bfZ),q(\bfU)) = \nonumber \\
& \; \; \; \; \int_{\bfU,\bfX} q(\bfU) q(\bfX|\bfZ) \log \frac{p(\bfU)p(\bfX)\int_\bfF p(\bfY | \bfF)p(\bfF|\bfU,\bfX)}{q(\bfU)q(\bfX|\bfZ)} \nonumber \\
&= \int_{\bfU,\bfX} q(\bfU) q(\bfX|\bfZ) \log \frac{p(\bfU)\int_\bfF p(\bfY | \bfF)p(\bfF|\bfU,\bfX)}{q(\bfU)} \nonumber \\
&- \KL{q(\bfX|\bfZ)}{p(\bfX)} \nonumber \\
&\coloneqq \hat{\F} - \KL{q(\bfX|\bfZ)}{p(\bfX)}.
\end{align}

At this point, our variational bound is similar to the one of equation \eqref{eq:boundSemiSupervised}, but the first term, here denoted as $\hat{\F}$,
refers to the expanded probability space and, thus, involves the inducing inputs and the additional variational
distribution $q(\bfU)$. Since the second term (the KL term) is tractable (because it only involves Gaussian distributions),
we are now going to focus on the $\hat{\F}$ term.
By breaking the logarithm again, we can further write this term as:
\begin{align}
\hat{\F} 
&= \int_{\bfU,\bfX} q(\bfU)q(\bfX|\bfZ) \log \left( \int_\bfF p(\bfY | \bfF) p(\bfF | \bfU, \bfX) \right) \nonumber \\
&- \KL{q(\bfU)}{p(\bfU)} \; \; \text{(A.1)} \label{eqApp:boundF1}.
\end{align}

We notice that we can make use of Jensen's inequality once more, because:
$$
\log \left( \int_\bfF p(\bfY | \bfF) p(\bfF | \bfU, \bfX) \right) \ge  \int_\bfF p(\bfF | \bfU, \bfX) \log p(\bfY | \bfF).
$$
This expectation is analytically tractable. Indeed, for a single dimension $\jd$, we can find this expectation as:
\begin{align*}
& \int_{\bff\d} p(\bff\d | \bfu\d, \bfX) \log p(\bfy\d | \bff\d)
=  \nonumber \\
& \qquad \log \gaussianDist{\dataVector_\outputIndex}{\mathbf{a}_\outputIndex}{\beta^{-1} \eye} - \frac{\beta}{2} \tr{\Kff } \nonumber \\
& \qquad +  \frac{\beta}{2} \tr{\Kuu^{-1} \Kuf\Kfu}, \label{eq:expectYgivenF}
\end{align*}
where $\bfK$ is the covariance matrix constructed by evaluating the covariance function on the training inputs $\mX$.
The full expression can be found by taking the appropriate product with respect to dimensions; indeed, since the joint probability factorises with respect to output dimensions $\jd$, then a bound to the logarithm of the marginal likelihood can be written as a sum over terms, where every term considers a single dimension $\jd$. Notice that to obtain this tractable bound we did not explicitly make the assumption of equation (\ref{eq:varDistr}) about the form of the variational distribution. However, this assumption is still made implicitly and the equivalence of the two derivations is rather instructive with respect to the effect of a variational constraint.

We also notice that in the above expression, the covariance matrix $\Kff$ is no longer inverted. Therefore, by writting the term $\hat{\F}$ in this form, 
we manage to obtain an expression which allows the uncertainty in $\bfX$ to be propagated through the GP mapping. 

It is possible to also obtain a ``tighter'' variational bound $\F(q(\bfX | \mXo)) \ge \F(q(\bfU),q(\bfX|\mXo))$ which does not depend on $q(\bfU)$.
To do so, we need to ``collect'' all terms that contain $p(\bfU)$ from equation (A.1) and find the stationary point
with respect to the distribution $q(\bfU)$ (by computing the gradient w.r.t $q(\bfU)$ and setting it to zero). By doing so, we are then able to replace $q(\bfU)$ with its optimal value back to the variational bound. \cite{BayesianGPLVM} further explain this trick.

\section{APPENDIX: MORE DETAILS FOR THE SEMI-DESCRIBED LEARNING EXPERIMENT}

\begin{figure*}[th]
\begin{center}
       \includegraphics[width=1\textwidth]{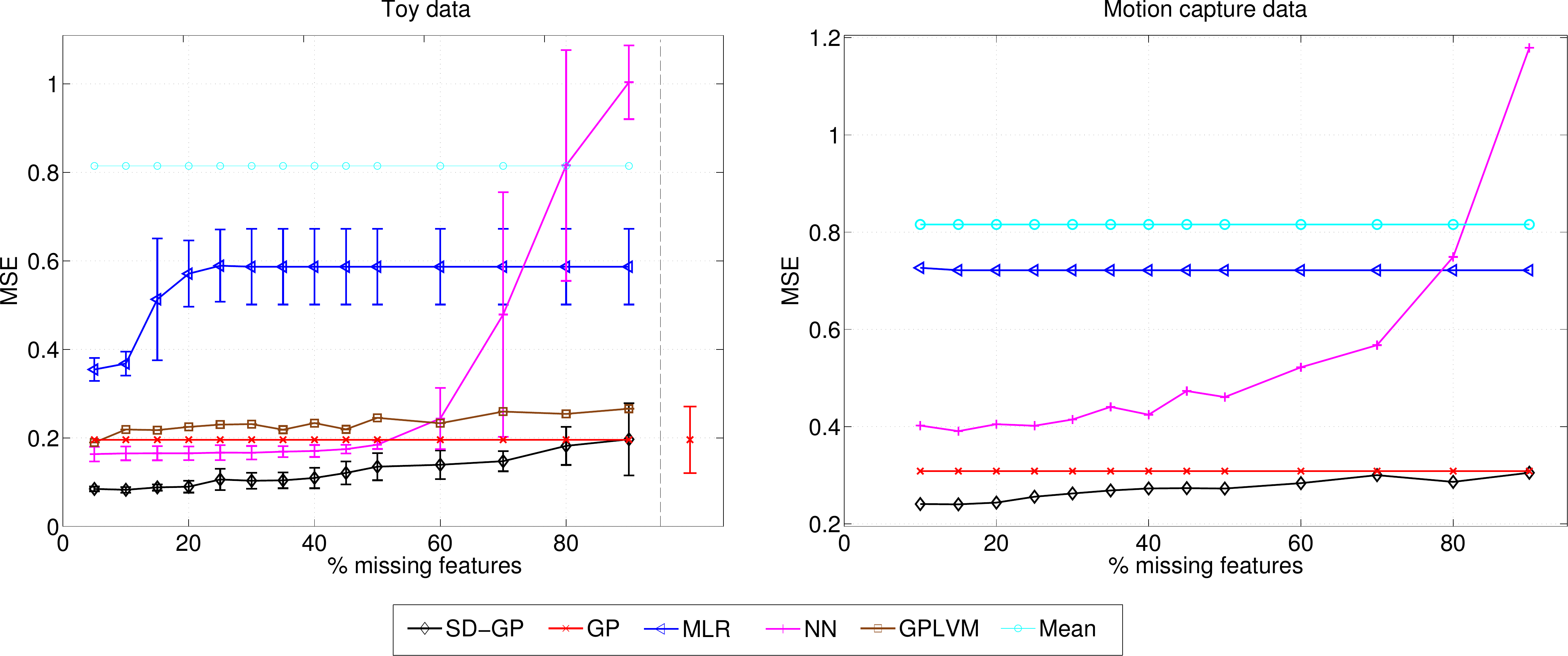}
\end{center}
\caption[Semi-described GP: partially missing inputs experiment.]{
MSE for predictions obtained by different methods on semi-described learning (full version of figure \ref{fig:semiSupervised}). Comparing our method (SD-GP), the standard GP method, multiple linear regression (MLR), nearest neighbour regression on the input space (NN), the data-imputation method based on GP-LVM and the mean predictor (mean). The results for simulated data are obtained from 4 trials. The GP method cannot handle partial observations, thus the uncertainty ($2\sigma$) is constant; for clarity, the errorbar is plotted separately on the right of the dashed vertical line (for nonsensical $x$ values). The GP-LVM method produced huge errorbars (about 3.5 times larger than thos of MLR), thus we don't plot them here, for clarity.}
\label{fig:semiSupervisedSuppl}
\end{figure*}

In Section \ref{subsec:semiDescribed} we looked at performing predictions with Gaussian processes trained from partially observed inputs. Our method (semi-described GP or SD-GP) was compared to other approaches in figure \ref{fig:semiSupervised}, but the limit in the $y-$axis was fixed to a smaller value to show the comparison with the standard GP method more clearly. For the same reason, methods which produced very large errors were omitted. In this appendix we show the full figure from all the obtained results -- figure \ref{fig:semiSupervisedSuppl}.

The conclusion drawn from figure \ref{fig:semiSupervisedSuppl} is that our method is very efficient in taking into account the extra, partially observed input set $\mXo^\unobservedSet$. This is true even if this extra set only has a small proportion of features observed. On the other hand, nearest neighbour runs into difficulties when real data are considered and, even worse, produces huge errors when more than 60\% of the features are missing in $\mXo^\unobservedSet$. Finally, the baseline which uses the standard GP-LVM as a means of imputing missing values produces bad results, in fact worse compared to if the extra set $\mXo^\unobservedSet$ is just ignored (\ie the GP baseline). This is because the baseline using GP-LVM treats the input space as single point estimates; 
 by not incorporating (and optimising jointly) the uncertainty for each input location, the model has no way of ignoring ``bad'' imputed values.

\section{APPENDIX: MORE DETAILS FOR THE AUTO-REGRESSIVE EXPERIMENT}

\begin{figure*}[t]
\begin{center}
   \includegraphics[width=1\textwidth]{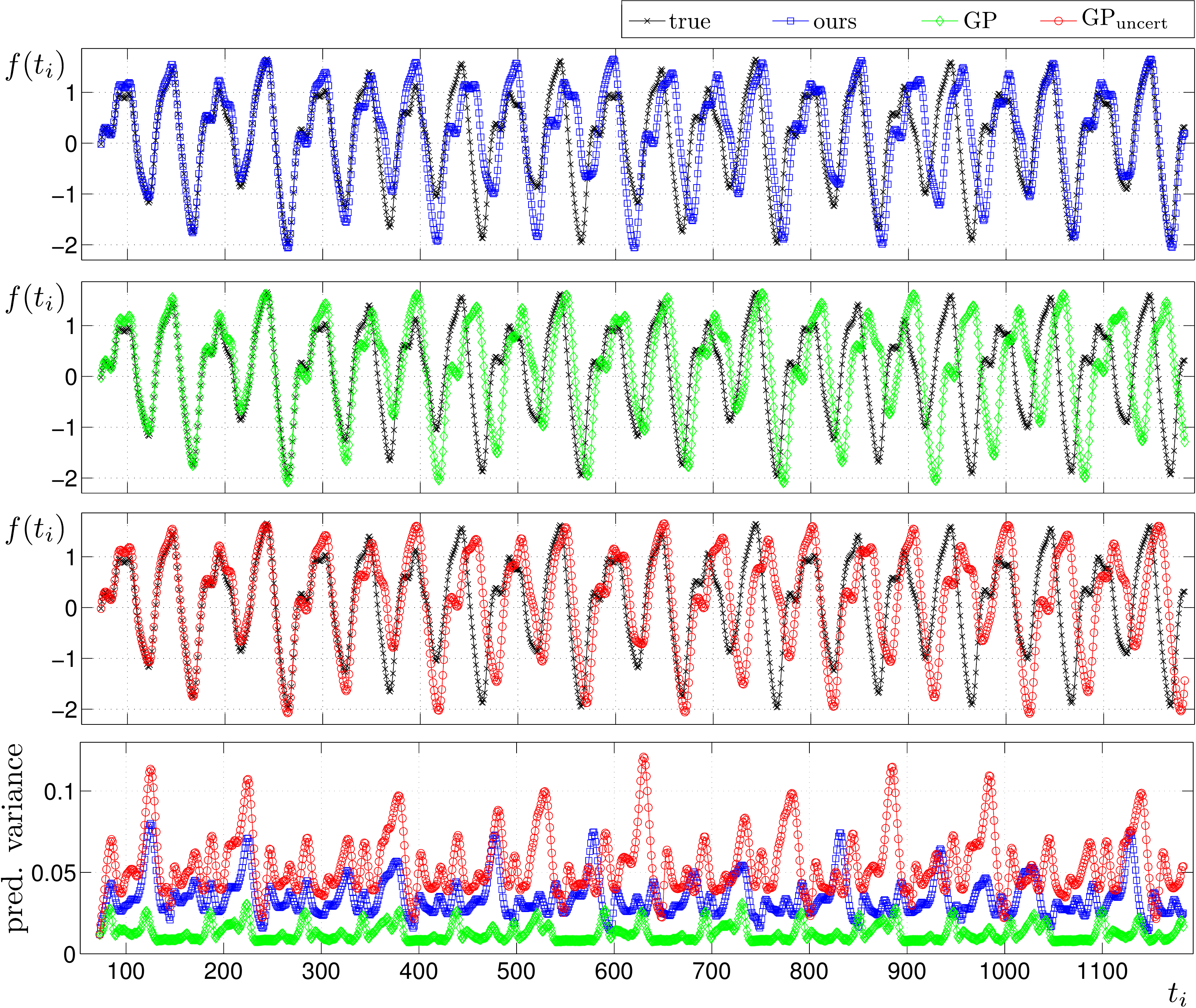}
\end{center}
\caption[The full picture in the extrapolation test.]{
The full predictions obtained by the competing methods for the chaotic time-series data. The top 3 plots show the values obtained in each predictive step for each of the compared methods; the plot on the bottom shows the corresponding predictive uncertainties ($2 \sigma$).
$\text{GP}_{\text{uncert}}$ refers to the basic (moment matching) method of \cite{Girard:uncertain01Compact} and the GP is the ``naive'' autoregressive GP which does not propagate uncertainties.
}
\label{fig:uncInputsExtrapolationMore}
\end{figure*}

This appendix refers to the auto-regressive Gaussian process model developed in Section \ref{subsec:autoregressive}. 
In figure \ref{fig:uncInputsExtrapolation} we showed the results from the last 310 steps of the iterative forecasting task. Here (figure \ref{fig:uncInputsExtrapolationMore}) we show the rest of the predictive sequence, obtained for extrapolating up until 1110 steps. The corresponding quantification of the error is shown in Table \ref{table:uncInputsError}.

\begin{table}[h]
  \caption{Mean squared and mean absolute error obtained when extrapolating in the chaotic time-series data. $\text{GP}_{\text{uncert}}$ refers to the basic (moment matching) method of \cite{Girard:uncertain01Compact} and the ``naive'' autoregressive GP approach is the one which does not propagate uncertainties.}
  \label{table:uncInputsError}
  \begin{center}
    \begin{tabular}{c||c|c}
      Method                                 & MAE              & MSE \\ 
      \hline \hline
      ours                            & $\mathbf{0.529}$ & $\mathbf{0.550}$ \\ \hline
      $\text{GP}_{\text{uncert}}$         & $0.700$          & $0.914$ \\ \hline
      ``naive'' GP approach & $0.799$          & $1.157$ \\
    \end{tabular}
  \end{center}
\end{table}

\section{APPENDIX: THE EFFECT OF $\nQ, \nD, \nN$ IN SEMI-DESCRIBED LEARNING\label{app:effectQDN}}
\begin{figure}[h]
\begin{center}
       \includegraphics[width=0.43\textwidth]{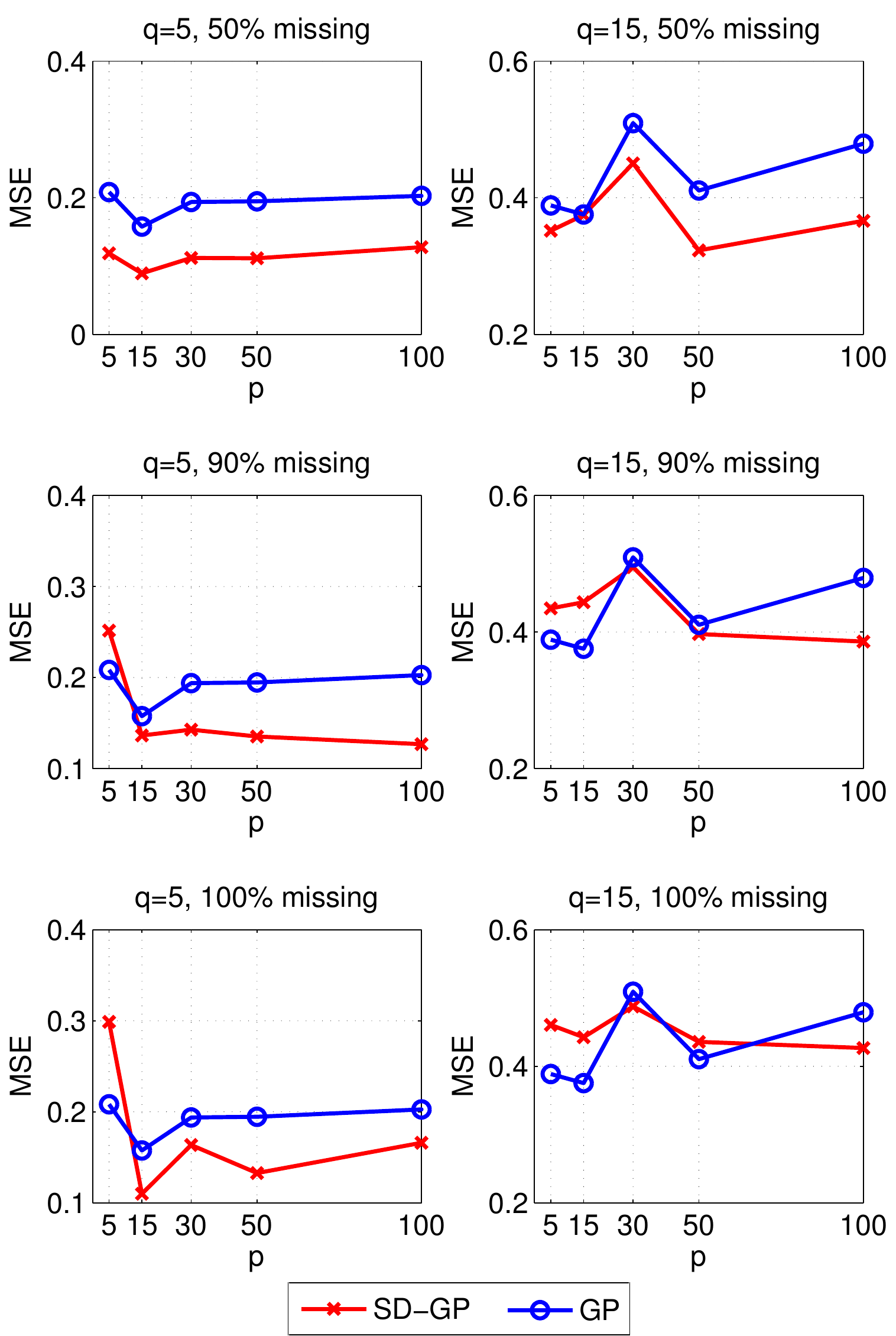}
\end{center}
\caption[Semi-described learning comparison for different dimensionalities.]{Comparison of our method (SD-GP) and the standard GP (which ignores $\mXo^\unobservedSet$) for different selections of $\nQ$, $\nD$ and percentage of missing features in $\mXo^\unobservedSet$.}

\label{fig:semiSupervisedPandQSuppl}
\end{figure}

As mentioned in Section \ref{subsec:semiDescribed}, we found that when $\nQ$ is large compared to $\nD$ and $\nN$, then the data imputation step of our algorithm can be problematic as the percentage of missing features in $\mXo^\unobservedSet$ approaches $ 100\%$. This is somehow a corner-case, but it still shows that the method is reliant on having some covariates available. To investigate further this issue we created simulated data as explained in Section \ref{subsec:semiDescribed}, but this time multiple datasets were generated with different input and output dimensions, $\nQ$ and $\nD$ respectively. In figure \ref{fig:semiSupervisedPandQSuppl} we show the comparison of SD-GP and the standard GP (which ignores $\mXo^\unobservedSet$) for different selections of $\nQ$, $\nD$ and percentage of missing features in $\mXo^\unobservedSet$.

The summary of this experiment is that:
\begin{itemize}
\item For the most usual scenarios, \ie when the percentage of features missing is not too high, SD-GP performs very well, but as $\nD$ and $\nN$ become small compared to $\nQ$, then the performance of the method seems to deteriorate.
\item Even if 100\% of the features are missing in $\mXo^\unobservedSet$, using our SD-GP can still be advantageous compared to using a standard GP. This is because SD-GP can utilise the extra information in the fully observed outputs, $\mY^\unobservedSet$, which correspond to the fully missing set $\mXo^\unobservedSet$. However, when the percentage of missing features is very large and the relative size of $\nD$ and $\nN$ is small compared to $\nQ$, then the method can produce worst results compared to the standard GP. 
\end{itemize}

To explain the challenge of handling missing values with SD-GP, consider that a separate variational parameter exists for every input, namely the parameters $\mu_{i,j}^\unobservedSet, S_{i,j}^\unobservedSet, \jn=1, ... \nN,\jq=1, ... , \nQ$ of step 7 in Algorithm \ref{algorithm:semiSupervised}. In the extreme cases mentioned in the previous paragraph, the number of variational parameters remains large but the available covariates to learn from are too few. This renders the optimisation of the parameters very difficult. 

\end{document}